%% file: acl2023.tex
\pdfoutput=1

\documentclass[11pt]{article}
\usepackage{overpic}
\usepackage{float}

\newif\ifcomment\commenttrue
\input{style/preamble}

\usepackage[]{acl2023}

\usepackage{xspace}
\usepackage{xcolor}
\usepackage[utf8]{inputenc}
\usepackage{pgfplots}
\usepackage{dsfont}
\DeclareUnicodeCharacter{2212}{−}
\usepgfplotslibrary{groupplots,dateplot}
\usetikzlibrary{patterns,shapes.arrows}
\pgfplotsset{compat=newest}

\usepackage{tikzscale}
\usepackage{relsize}
\usepackage{amsmath,amssymb}
\newcommand{\probP}{\text{I\kern-0.15em P}}

\usepackage{multirow, colortbl}

\usepackage{tabularx,booktabs}
\usepackage{makecell}

\usepackage[normalem]{ulem}
\useunder{\uline}{\ul}{}

\usepackage{graphicx}

\definecolor{ablation6}{HTML}{fcefed}
\definecolor{ablation_tie}{HTML}{fce3e1}

\definecolor{ablation5}{HTML}{fcd8d4}
\definecolor{ablation4}{HTML}{FBC3BC}
\definecolor{ablation3}{HTML}{F7A399}
\definecolor{ablation2}{HTML}{F38375}
\definecolor{ablation1}{HTML}{EF6351}

\definecolor{OliveGreen}{rgb}{0.05, 0.75, 0.24}
\definecolor{BrickRed}{rgb}{0.8, 0.25, 0.33}

\newcommand{\positive}[1]{\textcolor{OliveGreen}{#1}}
\newcommand{\negative}[1]{\textcolor{BrickRed}{#1}}

\usepackage[]{algpseudocode}
\usepackage[]{algorithm}
\usepackage{float}
\algtext*{EndFor}%
\algtext*{EndProcedure}%

\usepackage{booktabs}
\usepackage[normalem]{ulem}
\useunder{\uline}{\ul}{}

\usepackage{times}
\usepackage{latexsym}
\usepackage{adjustbox}

\usepackage[T1]{fontenc}
\usepackage{amsmath}
\usepackage{amssymb}
\usepackage{booktabs}
\usepackage{tikzscale}
\usepackage{amsmath}

\newcommand{\specialcellleft}[2][l]{%
\begin{tabular}[#1]{@{}l@{}}#2\end{tabular}}

\usepackage{multirow, colortbl}

\usepackage{tabularx,booktabs}
\usepackage{makecell}
\usepackage{multirow}
\usepackage{scalerel,xparse}

\usepackage{cleveref}
\usepackage{microtype}
\usepackage[most]{tcolorbox}

\usepackage{enumitem}
\crefformat{section}{\S#2#1#3}
\crefformat{subsection}{\S#2#1#3}
\crefformat{subsubsection}{\S#2#1#3}

\definecolor{bggray}{rgb}{0.95, 0.95, 0.95}
\usepackage[%
    framemethod=tikz,
    skipbelow=\topskip,
    skipabove=\topskip
]{mdframed}
\mdfsetup{%
    leftmargin=0pt,
    rightmargin=0pt,
    backgroundcolor=bggray,
    middlelinecolor=black,
    roundcorner=3
}

\definecolor{SkyBlue}{rgb}{0.53, 0.81, 0.92}

\newtcolorbox[
  list inside=prompt,
  auto counter,
  number within=section
]{prompt}[1][]{%
  enhanced,
  float*=t,                 
  colbacktitle=black!60,
  fonttitle=\small,
  coltitle=white,
  fontupper=\footnotesize,
  boxsep=4pt,
  left=0pt, right=0pt, top=0pt, bottom=0pt,
  boxrule=1pt,
  width=\textwidth,          
  enlarge left by=0mm,
  enlarge right by=0mm,
  listing only,
  listing options={
    basicstyle=\ttfamily\footnotesize,
    breaklines=true,
    breakatwhitespace=true,
    language=json
  },
  #1,
}

\newtcolorbox[
  list inside=trace,
  auto counter,
  number within=section
]{trace}[1][]{%
  enhanced,
  float*=t,
  colback=blue!5,             
  colbacktitle=blue!60!black, 
  colframe=blue!60!black,     
  fonttitle=\small,
  coltitle=white,
  fontupper=\footnotesize,
  boxsep=4pt,
  left=0pt, right=0pt, top=0pt, bottom=0pt,
  boxrule=1pt,
  width=\textwidth,
  enlarge left by=0mm,
  enlarge right by=0mm,
  listing only,
  listing options={
    basicstyle=\ttfamily\footnotesize,
    breaklines=true,
    breakatwhitespace=true,
    language=json
  },
  #1,
}

\definecolor{UMDred}{HTML}{ed1c24}

\definecolor{yellowcolor}{HTML}{ffc20e}
\definecolor{redcolor}{HTML}{e99999}
\definecolor{orangecolor}{HTML}{f6b26b}
\definecolor{yellowcolor}{HTML}{ffd966}
\definecolor{bluecolor}{HTML}{a0c5e8}
\definecolor{purplecolor}{HTML}{d9d2e9}

\usepackage{xcolor}
\definecolor{highlight}{HTML}{FFAE02}

\title{\raisebox{-0.1em}{\includegraphics[height=1em]{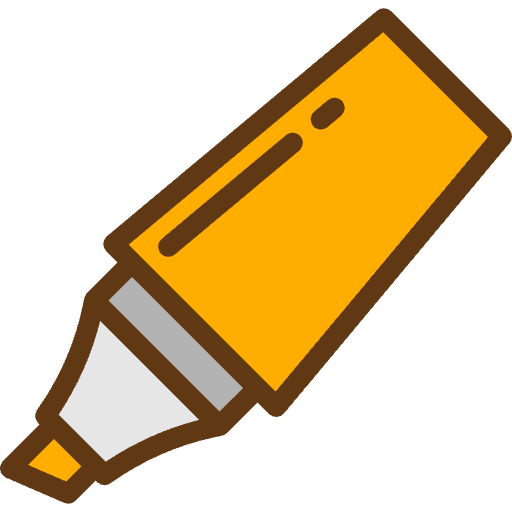}} \model{}: An Education-Inspired Toolkit for Highlighting\\Flaws in Multiple-Choice Benchmarks}

\newcommand{\authorSpacing}{0.2cm}
\author{
\textbf{Nishant Balepur}$^{1, 2}$\hspace{\authorSpacing}
\textbf{Bhavya Rajasekaran}$^{1}$ \hspace{\authorSpacing}
\textbf{Jane Oh}$^{1}$ \hspace{\authorSpacing}
\textbf{Michael Xie}$^{1}$ \hspace{\authorSpacing}
\textbf{Atrey Desai}$^{1}$ \\
\textbf{Vipul Gupta}$^{3}$ \hspace{\authorSpacing}
\textbf{Steven Moore}$^{4}$ \hspace{\authorSpacing}
\textbf{Eunsol Choi}$^{2}$ \hspace{\authorSpacing}
\textbf{Rachel Rudinger}$^{1}$ \hspace{\authorSpacing}
\textbf{Jordan Boyd-Graber}$^{1}$ \\[0.5em]
$^{1}$University of Maryland \hspace{0.3cm}
$^{2}$New York University \hspace{0.3cm}
$^{3}$Scale AI \hspace{0.3cm}
$^{4}$George Mason University\\[0.5em]
\texttt{nbalepur@umd.edu} \hspace{0.5em} \texttt{jbg@umiacs.umd.edu}
}

\begin{document}
\maketitle

\input{2026_arr_metabench/sections/00_abstract}

\input{2026_arr_metabench/sections/10_intro}

\input{2026_arr_metabench/sections/20_method}

\input{2026_arr_metabench/sections/30_judges}

\input{2026_arr_metabench/sections/40_results}

\input{2026_arr_metabench/sections/50_related_work}

\input{2026_arr_metabench/sections/60_conclusion}

\input{2026_arr_metabench/sections/70_limitation_ethics}

\bibliography{custom}
\bibliographystyle{acl_natbib}

\clearpage

\appendix
\input{2026_arr_metabench/sections/100_appendix}

\end{document}

%% file: style/preamble.tex
\usepackage[a-1b]{pdfx}

\usepackage{framed}
\usepackage{lmodern}
\usepackage{mdwlist}
\usepackage{siunitx}
\usepackage{latexsym}
\usepackage{colortbl}
\usepackage{xcolor}
\usepackage{nicefrac}
\usepackage{booktabs}
\usepackage{fnpct}
\usepackage{amsfonts}
\usepackage[T1]{fontenc}
\usepackage{bold-extra}
\usepackage{amsmath}
\usepackage{amssymb}
\usepackage{bm}
\usepackage{graphicx}
\usepackage{mathtools}
\usepackage{microtype}
\usepackage{multirow}
\usepackage{multicol}
\usepackage{xpatch}
\usepackage{latexsym,comment}
\usepackage[normalem]{ulem}

\newcommand*{\missingreference}{{\Huge \colorbox{red}{?reference?}}}
\newcommand*{\missingcitation}{{\Huge \colorbox{red}{?citation?}}}

\makeatletter
\xpatchcmd{\@setref}{\bfseries}{\missingreference}{}{}
\def\@citex[#1]#2{\leavevmode
    \let\@citea\@empty
    \@cite{\@for\@citeb:=#2\do
        {\@citea\def\@citea{,\penalty\@m\ }%
            \edef\@citeb{\expandafter\@firstofone\@citeb\@empty}%
            \if@filesw\immediate\write\@auxout{\string\citation{\@citeb}}\fi
            \@ifundefined{b@\@citeb}{\hbox{\reset@font\missingcitation}%
                \G@refundefinedtrue
                \@latex@warning
                {Citation `\@citeb' on page \thepage \space undefined}}%
            {\@cite@ofmt{\csname b@\@citeb\endcsname}}}}{#1}}
\makeatother

\newcommand{\model}[0]{BenchMarker\xspace}

\newcommand{\mm}[0]{\textsc{llm}\xspace}
\newcommand{\mcqa}{\textsc{mcqa}\xspace}

\newcommand{\mcq}{\textsc{mcq}\xspace}

\newcommand{\gem}[1]{\mbox{\textsc{gem}}}
\newcommand{\abr}[1]{\textsc{#1}\xspace}

\newcommand{\hidetext}[1]{}
\newcommand{\ignore}[1]{}

\ifcomment
    \newcommand{\pinaforecomment}[3]{\colorbox{#1}{\parbox{.8\linewidth}{#2: #3}}}

    \newcommand{\prtodo}[1]{\pinaforecomment{lightblue}{pr}{#1}}
    \newcommand{\prtodoi}[1]{\pinaforecomment{lightblue}{pr}{#1}}
\else
    \newcommand{\pinaforecomment}[3]{}
    \newcommand{\prtodo}[1]{}
    \newcommand{\prtodoi}[1]{}
\fi

\newcommand{\smallurl}[1]{ \begin{tiny}\url{#1}\end{tiny}}

\definecolor{lightblue}{HTML}{3cc7ea}
\definecolor{CUgold}{HTML}{CFB87C}
\definecolor{grey}{rgb}{0.95,0.95,0.95}
\definecolor{ceil}{rgb}{0.57, 0.63, 0.81}
\definecolor{UMDred}{HTML}{ed1c24}
\definecolor{UMDyellow}{HTML}{ffc20e}

\newcommand{\nlp}[0]{\abr{nlp}}


%% file: 2026_arr_metabench/sections/00_abstract.tex
\begin{abstract} {
Multiple-choice question answering (\mcqa{}) is standard in~\nlp{}, but benchmarks lack rigorous quality control. 
We~present \model{}, an education-inspired toolkit using \mm{} judges to flag three common \mcq{} flaws:
1) \textit{contamination}---items appearing exactly online;
2) \textit{shortcuts}---cues in the choices that enable guessing; 
and 3) \textit{writing errors}---structural/grammatical issues based on a 19-rule education rubric.
We validate \model{} with human annotations, then run the tool to audit 12 benchmarks, revealing:
1) flaws persist in \mcqa{} benchmarks, especially automatically-made and crowdsourced data---we detect 47\% of TruthfulQA appears online and 100\% of HellaSwag violates multiple writing rules;
2) contaminated \mcq{}s tend to inflate accuracy, while writing errors tend to lower it and change rankings beyond random;
and 3) prior benchmark repairs address their targeted 
issues (i.e., lowering accuracy with \mm{}-written distractors), but inadvertently add~new flaws (i.e. implausible distractors, many correct answers).
Overall, flaws in \mcq{}s degrade \nlp evaluation, but education research offers a path forward. We release \model{} to bridge the fields and improve \mcqa{} benchmark design.\footnote{Our code and data are available at: \url{https://github.com/nbalepur/BenchMarker}}
}
\end{abstract}

%% file: 2026_arr_metabench/sections/10_intro.tex
\input{figures/intro}

\section{Grading Benchmarks like Educators} \label{section:intro}

Progress in \nlp{} relies on benchmarks \cite{voorhees2001trec} that are largely multiple-choice question answering (\mcqa{}) tasks, where models must pick the answer to a question from input choices \cite{robinson2023leveraging}.
\mcq{}s test whether models can understand the~question, recall related facts, and use said knowledge to deduce the answer \cite{richardson2013mctest}, but only do so reliably when~error-free.
\mcq{} flaws add noise unrelated to these skills, undermining their construct validity \cite{smith2005construct}.

\mcqa{} datasets are notoriously~flawed: \mcq{}s exist in \mm{} training data \cite{deng2023investigating}, have exploitable shortcuts \cite{gupta2024improving},~and are rife with writing~issues \cite[e.g., poor grammar]{Palta2024PlausiblyPQ}.
\textit{Educators} detect, cull, and remedy these errors when writing \mcq{}s to ensure students 
rarely face them \cite{campbell2011write}, but while \nlp{} \textit{researchers} value \mcqa{} for its similarity to human testing \cite{zhuangposition}, they rarely adopt~education's standards: for 39 surveyed \mcqa{} datasets, 23\% report no quality control (Appendix~\ref{appendix:survey}).

To address this, prior work has devoted extensive resources to ``correct'' \mcqa{} benchmarks~\cite{wang2024mmlu, chizhov2025hellaswagvaliditycommonsensereasoning}.
However, they do not propose reusable metrics to check whether rewritten \mcqa{} benchmarks truly reduce errors.


We present \textbf{\model{}}, a toolkit for detecting \mcq{} errors based on~three metrics educators value (Fig~\ref{fig:main}):
\textit{1) contamination}---whether \mcq{}s appear online, a proxy for dataset leakage, similar to ensuring students cannot cheat on exams \cite[\cref{subsection:contamination}]{ConceptInventory};
\textit{2) shortcuts}---whether \mcq{}s have~shallow cues that let strong~\mm{}s answer without the question, like students guessing via partial knowledge \cite[\cref{subsection:shortcuts}]{lau2011guessing}; and
\textit{3) writing errors}---grammar and structure violations on~a 19-rule rubric of educational \mcq{} writing guidelines \cite[\cref{subsection:writing_flaws}]{costello2018evaluation}.
We distil these insights from decades of education research via \nlp{} advances in \mm{}-as-a-judge \cite{zheng2023judging}, validated across 23 models, six web search \textsc{api}s, 13 \mcqa{} datasets, and 8042 expert judgments~(\cref{section:setup}).

\model{} audits 12 \mcqa{} datasets and predicts pervasive issues, especially~with \mcq{}s from \nlp{} annotation protocols, versus exams that educators design: 47\% of TruthfulQA appears online, 21\% of ScholarIQA 
has shortcuts, and HellaSwag always violates at least two writing rules (\cref{subsection:eval_benchmark}). 
Educational theory informs that these flaws degrade \mcqa{} \cite{cronbach1955construct}, so~we empirically assess their impact on \mm{} evaluation.
\mm{}s have higher accuracy on contaminated data splits and lower accuracy on splits 
with multiple writing errors; discarding \mcq{}s with writing errors shifts \mm{} rankings beyond random  (\cref{subsection:rankings}), muddying how researchers select models to train and deploy.
The writing errors from \nlp{} and educational assessments overlap---unclear wording and distractor quality---motivating these fields to collaborate and jointly advance evaluations and student testing (\cref{subsection:which_writing_flaws}).
Lastly, while past benchmark fixes resolve targeted issues, they can add new ones: MMLU-Pro uses \mm{}-written distractors to lower accuracy, but this produces faulty distractors and more than one correct answer (\cref{subsection:fixes}).
Thus, \mcq{} correction requires iterative refinement, which \model{} can support via repeatable, automated scoring runs.


%
%

The quality of \mcqa{} benchmarks is often overlooked, but this erodes evaluation validity.
Education research can guide the design quality control tools like \model{} and offer lessons to \nlp{} from student testing, so we champion more collaborations between these fields.
We contribute:

\begin{enumerate}[leftmargin=1em, labelsep=0.5em]
    \item \model{}, a toolkit to predict contamination, shortcuts, and writing flaws in \mcq{}s.~We~wrap it in InspectAI, a library with judge logs, standard prompts, and a \textsc{ui} to track runs (Appendix~\ref{appendix:inspect}).

    \item A rigorous audit of 12 \nlp{} benchmarks, showing \mcqa{} flaws are pervasive and impact \mm{} accuracy and rankings, especially writing errors on crowdsourced and automatically written \mcq{}s.

    \item In-depth analysis on common writing flaws in \nlp{}'s and education's \mcq{}s, and evidence that prior solutions do not fully address these issues.

    \item A validation dataset of 8042 human judgments to test \mm{} judge reliability in \mcq{} flaw detection.
\end{enumerate}


%% file: figures/intro.tex
\begin{figure*}
    \centering
    \begin{overpic}[width=\linewidth]{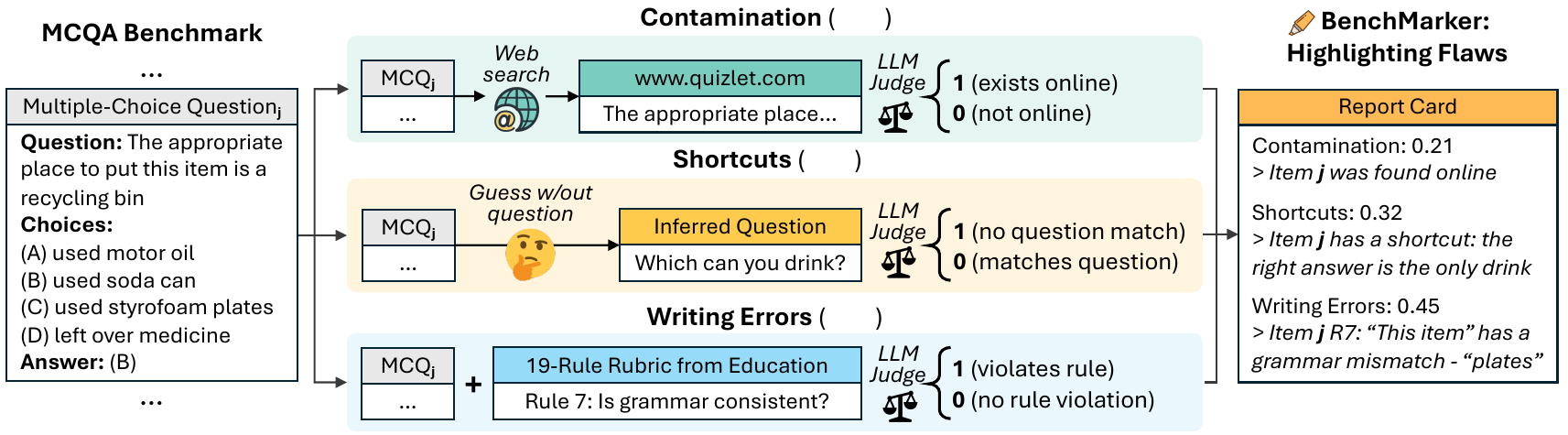}
    \put(53.3, 26){\scalebox{0.75}{\cref{subsection:contamination}}}
    \put(51.3, 17){\scalebox{0.75}{\cref{subsection:shortcuts}}}
    \put(52.7, 6.9){\scalebox{0.75}{\cref{subsection:writing_flaws}}}
    \end{overpic}
    \caption{\textbf{\model{}} scores \mcqa{} benchmark items across three axes with \mm{} judges: 1) \textit{contamination}---whether the~item appears on the Internet; 2) \textit{shortcuts}---whether models can use shallow shortcuts in choices to solve the item without the question; and 3) \textit{writing errors}---grammatical and structural issues based on a 19-rule rubric derived from education research. We aggregate scores to audit datasets and return judge feedback on items.}
    \label{fig:main}
\end{figure*}

%% file: 2026_arr_metabench/sections/20_method.tex
\section{\model{}: Marking \mcqa{} Flaws} \label{section:method}

A multiple-choice question (\mcq{}) has a question stem $q$ and choices $\mathcal{C}$.
Test-takers must pick the best answer $a \in \mathcal{C}$ and avoid distractors $\mathcal{D} = \mathcal{C} \setminus \{ a \}$.
\mcq{}s assess if test-takers~can understand questions, recall facts, and deduce~choice correctness \cite{simkin2005multiple}, but issues in \mcqa{} datasets undermine this goal~\cite{chizhov2025hellaswagvaliditycommonsensereasoning}.
We thus introduce \textbf{\model{}} (Figure~\ref{fig:main}), which builds upon education research to detect benchmark flaws.

We design \model{} to detect flaws \textit{within} \mcq{}s, leaving global issues like saturation \cite{vania2021comparing} and diversity \cite{uzunoglu2025flaw} as future work. 
For our blueprint, we review \mcqa{} work~in education and \nlp{} \cite{haladyna2002review, balepur-etal-2025-best}, informing us of three key issues: 
contamination (\cref{subsection:contamination}), shortcuts (\cref{subsection:shortcuts}), and writing errors~(\cref{subsection:writing_flaws}).
We use \mm{} judges to~predict each issue, later shown to agree with humans (\cref{section:setup}). 

\subsection{Contamination Detection} \label{subsection:contamination}

Educators typically write \mcq{}s that students have not seen before, testing the~transfer of learning beyond rote memorization  \cite{roediger2011critical}.
Equivalently, \nlp{} ensures that training and testing dataset splits do not overlap \cite{larson1931shrinkage}, 
assessing model generalization over memorization.

To score whether \nlp{} systems can easily memorize an \mcq{}, we check whether it exists online---a practice educators use \cite{ConceptInventory}. 
The~internet is a reasonable proxy for pre-training\footnote{Appendix~\ref{appendix:compare_contamination} compares to other contamination methods.} \cite{li-etal-2024-open-source}---such data is not fully disclosed~but likely on the web \cite{soldaini2024dolma}---and tests whether often-benchmarked retrieval-augmented \mm{}s \cite{bragg2025astabench} can cheat by finding the \mcq{} and echoing its answer \cite{paleka2025evaluating}.



We adapt \citet{li-etal-2024-open-source}'s web search check: 1) query a web \textsc{api} with question stem $q$ and gold answer $a$ as input; and 2) prompt an \mm{} to judge whether the item is exactly/nearly-exactly in search results, meaning it is contaminated.
We do not flag \mcq{}s when the web pages have only the \textit{knowledge} linked to the answer (i.e. no explicit \mcqa{}~format), since this would not lead to \textit{exact} memorization.

\input{data/datasets.tex}

\subsection{Shortcuts Detection} \label{subsection:shortcuts}

\mcqa{} is easy to score, as test-takers choose from a list, but this can enable informed guessing \cite{frary1988formula}:
partial knowledge students can exploit shortcuts in choices via meta-strategies to answer \mcq{}s \cite[e.g., pick the ``odd one out'', pattern matching]{lau2011guessing}, subverting the \mcq{}'s validity.
Likewise, \nlp{} models reach above-random \mcqa{} accuracy when answering using choices alone \cite{richardson2020does}, spurring concerns that \mcqa{} benchmarks have superficial cues that fail to assess true model abilities \cite{chandak2025answer}.

A standard way to identify shortcuts is \emph{partial-input studies}
\cite{poliak2018hypothesis}: isolating~\mcq{}s that \mm{}s can solve with just the choices.
Choices-only accuracy is often equated with shortcuts \cite{chandak2025answer}, but this can be problematic or benign.
For example, in Figure~\ref{fig:main} (middle), when answering with the choices alone, the \mm{} can guess which choice is the ``odd one out'' via shallow cues (\textit{``Soda'' is the only beverage in the list, so it must be right}), or more strategically infer the original question (\textit{Styrofoam plates, motor oil, and medicine are non-degradable, so the question likely links to recycling and ``soda can''  is right}).
The former is a concerning shortcut that lets models bypass the \mcq{}, undermining benchmark validity, but the latter is less problematic, since the model effectively answers the original \mcq{} as intended.


To draw this line and see if models can shortcut the \mcq{} in problematic ways, we prompt \mm{}s~to ``think step by step'' \cite{wei2022chain} to answer with the choices and then guess the original question, following \citet{Balepur2024ArtifactsOA}.
An \mm{} judge then evaluates whether the inferred question is semantically equivalent to the true question.
To limit model variance when answering without the question, we take a majority vote of three \mm{}s with high choices-only accuracy from \citet{balepur2025test}: GPT-5, Gemini 2.5 Pro, and Claude 4.5 Sonnet.\footnote{We report per-model shortcut detection in Appendix~\ref{appendix:shortcut_ablation}.}
We predict the \mcq{} as having shortcuts if: 1) \mm{}s solve the \mcq{} with just choices; and 2) the inferred question does not match the original, indicating that the \mcq{} has exploitable shortcuts. 

\subsection{Writing Error Detection} \label{subsection:writing_flaws}

\mcq{}s lose validity if authors add
structural, semantic, and grammatical errors while writing \mcq{}s, as they render the item misleading or unanswerable \cite{haladyna2002review}.
Analogously, if \nlp{} models fail here, we cannot discern whether they lacked skills \mcqa{} tests or misinterpreted poor writing.

Educators have long recognized such errors and have thus curated rubrics to rigorously assess \mcq{}s \cite{haladyna1989taxonomy}.
Inspired by this, we combine them with \mm{} judges \cite{gunjal2025rubrics}, which can follow rules to score~outputs with high human agreement \cite{kim2024prometheus}.
We take the 19-rule Item-Writing Flaws rubric from \citet{tarrant2006frequency}, which avoids subjective rules in others \cite[e.g., ``\textit{avoid trivial material}'']{haladyna1989taxonomy} and has been used for over two decades across domains in education---mostly higher education \cite{schmucker2025impact}.
We prompt \mm{}s to judge which rules an \mcq{} breaks, via~19 prompts with each~flaw~name, definition, and six examples (three flawed \mcq{}s, three flawless \mcq{}s).

The 19 rules ensure \mcq{}s are clear (e.g., no ambiguous terms like ``mostly'', state questions clearly/concisely in the stem), adhere to \mcqa{}'s format (e.g.,~have~one right answer, make distractors plausible), curb giveaways (e.g., the stem and choices must be grammatically consistent), and are not misleading (e.g., sort numerical options, avoid ``none of the above''); Table~\ref{table:writing_flaw_rules} has the full list.
Some rules may not always apply---work uses longer question stems to test long-context abilities \cite{dua2019drop} or ``none of the above'' for abstention \cite{elhady-etal-2025-wicked}---but we use all 19, since our audited benchmarks are more general (\cref{section:results}).
We eventually reveal the most pervasive issues are clarity and distractor quality (\cref{subsection:which_writing_flaws}), which apply to all \mcqa{} datasets.


%% file: data/datasets.tex
\begin{table*}[]
\small
\centering
\setlength{\tabcolsep}{3.2pt}
\begin{tabular}{@{}lccc@{}}
\toprule
Dataset                                            & Domain                          & Difficulty  & Creation Strategy    \\ \midrule
Algebra Question Answering \cite[AQuA]{zhong2024agieval}                  & Math                            & Graduate    & Student Exams                      \\
AI2 Reasoning Challenge \cite[ARC]{clark2018think}                      & Science                         & Elementary  & Student Exams                      \\
CommonsenseQA \cite[CQA]{talmor2018commonsenseqa}                                & Commonsense                     & General     & Crowdworkers                      \\
HellaSwag \cite[HS]{zellers2019hellaswag}                                     & Commonsense & General     & Model Generated \\
Multitask Language Understanding \cite[MMLU]{hendrycks2020measuring}    & Multi-Subject                   & College     & Student Exams                     \\
Open Book Question Answering \cite[OBQA]{mihaylov2018can}                & Science                         & Elementary  & Crowdworkers                     \\
Physical Interaction Question Answering \cite[PIQA]{bisk2020piqa}     & Commonsense                     & General     & Crowdworkers                      \\ 
Question Answering via Sentence Composition \cite[QASC]{khot2020qasc} & Science                         & Elementary  & Crowdworkers                     \\
Scholastic Aptitude Test \cite[SAT]{zhong2024agieval}                     & Math                            & High School & Student Exams                      \\
Social Intelligence Question Answering \cite[SIQA]{sap2019socialiqa}      & Commonsense                     & General     & Crowdworkers                     \\
Super Google Proof Question Answering \cite[SGPQA]{du2025supergpqa}      & Multi-Subject                   & Graduate    & Expert+Model                      \\
Truthful Question Answering \cite[TQA]{lin2021truthfulqa}                  & Commonsense                       & General     & Author-Written                     \\ \bottomrule
\end{tabular}
\caption{\small The \mcqa{} benchmarks we mainly explore. We audit \mcq{}s across varied domains, difficulties, and creation strategies. \label{table:datasets}}
\end{table*}

%% file: 2026_arr_metabench/sections/30_judges.tex
\section{Validating \model{}'s Reliability} \label{section:setup}

Before using \model{}, we ensure it reliably predicts flaws.
This section outlines datasets~(\cref{subsection:datasets}), annotations (\cref{subsection:annotations}), and baselines (\cref{subsection:baselines}) to confirm \model{} agrees with human judgments (\cref{subsection:judges}).

\subsection{Collecting \mcq{}s for Human Review} \label{subsection:datasets}

We collect \mcq{}s to assess \model{} from 12 \mcqa{} datasets of varied domains, difficulties, and creation strategies for dataset design (Table~\ref{table:datasets}).
We could randomly sample \mcq{}s, but this might omit rare flaws.
We instead run \model{} via GPT-5 as the judge on each benchmark's training set, then sample with stratification~\cite{cochran1977sampling,  zouhar2025select}: for each metric, we sample up to ten \mcq{}s per dataset where GPT predicts the item is flawed and up to ten GPT predicts are not flawed.
This sampling better surfaces positive and negative cases, ensuring we have both labels for each metric.

\model{} is tailored for \nlp{} datasets, but our scores parallel what educators value when writing \mcq{}s \cite{haladyna2002review}, so we also test how well \mm{}s judge \mcq{}s designed for students.
We use \citet{costello2018future}'s 4123 labels on~the 19 writing rules from \cref{subsection:writing_flaws} on higher education exams in computer science, humanities, health sciences, psychology and math.
This yields in-domain (for \nlp{})~and out-of-domain (for students) data for writing error detection.
In total, we gather 229~\mcq{}s to label for contamination, 271 to label for shortcuts, 3419 to label for writing errors on \nlp{} data, and 4123 human-written \mcq{}s with existing writing error labels, resulting in 8042 entries for validation.

\input{data/judge_shortcuts_and_writing}

\subsection{Human Annotations on \mcq{}s} \label{subsection:annotations}

We now collect human annotations on our \mcq{}s~to validate judges.
For contamination, annotators~look for~the \mcq{} across four search engines (Google, Bing, DuckDuckGo, and Brave), labeling it as $1$ (flawed) if it exists exactly in any web page, $0$ (not flawed) otherwise.
For shortcuts, annotators compare GPT's inferred~question from just the choices to the original question stem.
They label the item as $1$ if the two do not semantically match---not testing the same concepts---following \citet{Balepur2024ArtifactsOA}, and $0$ otherwise.
For writing errors, annotators see a rule from the 19-item rubric and mark whether the \mcq{} violates~it~via its definition and examples, following \citet{moore2023crowdsourcing}.

We do not use crowdworkers to rate \mcq{} quality, since \citet{moore2023crowdsourcing} show they have 25\% disagreement with experts.
Thus, we use a protocol based on qualitative coding in \textsc{hci} \cite{bingham2023data}: Author A---an English-speaking \textsc{cs} graduate researcher with papers and annotation experience in \mcqa{}---labels each \mcq{}, and Author~B with the same background labels 50 random items for each metric (50 labels for each writing rule, for $19 \cdot 50 = 950$ in all) to evaluate reliability. They have >80\% agreement per metric (Appendix~\ref{appendix:annotation}).

\subsection{Baseline and Metric Selection} \label{subsection:baselines}

We now assess how well \mm{} judge predictions ($\hat{l}$) match humans ($l$) on our validation set.
For shortcut/writing errors, we use~23 open/closed \mm{}s across seven families:
1) Gemini 2.5 \cite[Lite, Flash, Pro]{comanici2025gemini}; 
2) GPT-5 \cite[Nano, Mini, Base]{openai_gpt5_system_card_2025}; 
3) Claude 4.5 \cite[Haiku, Sonnet]{anthropic2025claude_sonnet_4_5}; 
4) Command \cite[R, R+]{cohere2024command_r}; 
5) Qwen-3 \cite[0.6, 1.7, 4, 8, 14, 32B]{yang2025qwen3};
6) Gemma-3 \cite[4, 12, 7B]{team2025gemma}; and
7) LLaMA-3 \cite[1, 3, 8, 70B]{dubey2024llama}.
We prompt \mm{}s with default parameters and request a \textsc{json} with a prediction and explanation.

We also assess the Scalable Automatic Question Usability Evaluation Toolkit \cite[\textsc{saquet}]{moore2024automatic} for writing errors, which uses heuristics and GPT-5 judges to detect the 19 writing errors we study, but optimized on the out-of-domain validation set of student exams (\cref{subsection:datasets}).
Contamination~detection relies on web search quality (\cref{subsection:contamination}), so we only use our three best \mm{}s (Gemini Pro, GPT-5, Sonnet), and focus on testing six web search \textsc{api}s: Google, Bing, Perplexity, Exa, Tavily, and Serper. 

We report \textit{accuracy} (the proportion where $\hat{l}=l$), \textit{F1 Score} (harmonic mean of precision/recall, measuring how well $\hat{l}$ predicts~$l$ with class imbalance), and \textit{Cohen's~$\kappa$} \cite[non-random agreement of $\hat{l}$ and~$l$]{cohen1960coefficient}.
To interpret scores, we add trivial baselines for each flaw type: randomly (50/50) predict $0$ (not flawed) or~$1$ (flawed), always predict $0$, and always predict $1$.
Beating these baselines gives confidence that judges informatively detect flaws.

\input{data/judge_contamination}

\subsection{\model{} Agrees with Human Judges} \label{subsection:judges}

We now evaluate how well judges agree with humans. 
In contamination, Google is the best search \textsc{api}, and GPT-5 surpasses Gemini and Claude at using its web pages (Table~\ref{table:judge_contamination}).
F1 is similar to the ``Always Flawed'' baseline, but accuracy/Cohen's~$\kappa$ are much higher, and Google+GPT-5's low F1 mainly stems from~low recall; precision is $0.86$.
Manual analysis shows Google's \textsc{api} gives a subset of public search engine results---our annotations flag contaminated \mcq{}s \textsc{api}s may not surface---so we still deem our contamination detection strong.

In shortcuts/writing errors, many \mm{}s beat trivial baselines (Table~\ref{table:combined_shortcuts_writing}); the best models are closed-source (GPT-5, Gemini Pro), but open-weight~ones compete (Command R, Qwen 32B), so future work can study tuning smaller \mm{}s to close this gap for efficiency/reproducibility.
GPT-5 has higher~mean F1/Cohen's $\kappa$ than \textsc{saquet} on in-domain and out-of-domain data, making \model{} state-of-the-art at detecting writing flaws versus existing tools.
\model{}'s competitive scores on educator-written exams suggest it could also help educators. 

\input{figures/dataset_scores}
\input{data/accuracy_impact}

We want \model{} to be useful across \mcqa{} datasets, so we also report its generalization.
For the best \mm{}s in Tables~\ref{table:combined_shortcuts_writing} and \ref{table:judge_contamination}, accuracy's standard deviation is $0.07$ for shortcuts, $0.15$ for contamination, and $0.06$ for writing errors; we later reveal writing errors are crucial for \nlp{} to~fix (\cref{subsection:impact}) and encouragingly, standard deviation is low.

\subsection{Recommendation: \mm{}s in \model{}} \label{subsection:recommendation}

Overall, \mm{}s predict \mcq{} contamination, shortcuts, and each of the 19 writing errors with Cohen's $\kappa$ and accuracy matching standard \mm{} judge protocols \cite{zheng2023judging, bavaresco-etal-2025-llms}.
When using \model{} for the rest of our analyses (\cref{section:results}), we use the \mm{}s with the best Cohen's $\kappa$ in Tables~\ref{table:combined_shortcuts_writing} and \ref{table:judge_contamination} for the three flaw types.
For writing errors, we pick the \mm{} with the best Cohen's $\kappa$ for each of the 19 errors types (Appendix~\ref{appendix:implementation}).

Most of \model{}'s cost comes from writing flaw detection, which uses 19 \mm{} judges per item.
For researchers with limited resources, Gemini-2.5 Pro is likely too expensive; we recommend Gemini-2.5 Flash, which is only $0.05$ below Gemini-2.5 Pro in Cohen's $\kappa$ and is $\sim\frac{1}{4}$ of the cost.\footnote{https://ai.google.dev/gemini-api/docs/pricing}
We provide a more detailed cost breakdown in Appendix~\ref{appendix:cost}.

Finally, for researchers without access to closed-source \mm{}s, we recommend Cohere Command-R for writing flaw detection; it has the best Cohen's $\kappa$ out of all open-weight \mm{}s.
In Appendix~\ref{appendix:open_weight}, we run follow-up experiments on judge ensembling, confidence calibration, and writing flaw detection failures to further support this recommendation.



%% file: data/judge_shortcuts_and_writing.tex
\newcommand{\notsig}{\ensuremath{*}}

\begin{table*}[]
\small
\centering
\setlength{\tabcolsep}{3.5pt}
\renewcommand{\arraystretch}{0.7}
\begin{tabular}{@{}lccc|ccc|ccc@{}}
\multicolumn{1}{c}{} 
  & \multicolumn{3}{c}{\textit{Shortcuts}} 
  & \multicolumn{3}{c}{\textit{Writing (In Domain, NLP)}}  
  & \multicolumn{3}{c}{\textit{Writing (Out of Domain, Human)}} \\
\toprule
Method
  & Accuracy & F1 Score & Cohen's $\kappa$
  & Accuracy & F1 Score & Cohen's $\kappa$
  & Accuracy & F1 Score & Cohen's $\kappa$ \\ \midrule
Gemini 2.5 Lite         & 0.76 & 0.74 & 0.51 & 0.68 & 0.56 & 0.35 & 0.71 & 0.28 & 0.18 \\
Gemini 2.5 Flash        & 0.69 & 0.68 & 0.41 & 0.79\notsig & 0.62 & 0.47 & 0.85 & 0.38 & 0.31 \\
Gemini 2.5 Pro          & 0.70 & 0.69 & 0.43 & \textbf{0.82\notsig} & \textbf{0.66} & \textbf{0.53} & 0.86 & 0.39 & 0.33 \\
GPT-5 Nano              & 0.68 & 0.65 & 0.38 & 0.74 & 0.39 & 0.22 & 0.88 & 0.21 & 0.14 \\
GPT-5 Mini              & 0.77 & 0.72 & 0.53 & 0.75 & 0.55 & 0.38 & 0.84 & 0.32 & 0.25 \\
GPT-5                   & \textbf{0.82\notsig} & \textbf{0.75} & \textbf{0.61} & 0.81 \notsig & 0.63 & 0.50 & 0.87 & 0.37 & 0.30 \\
Claude 4.5 Haiku        & 0.78\notsig & 0.73 & 0.55 & 0.72 & 0.58 & 0.38 & 0.76 & 0.31 & 0.22 \\
Claude 4.5 Sonnet       & 0.81\notsig & 0.75 & 0.59 & 0.79\notsig & 0.63 & 0.48 & 0.83 & 0.36 & 0.28 \\
Command R               & 0.74 & 0.72 & 0.50 & 0.77 & 0.53 & 0.38 & 0.89 & 0.37 & 0.31 \\
Command R+              & 0.72 & 0.70 & 0.46 & 0.76 & 0.53 & 0.36 & 0.88 & 0.36 & 0.30 \\
Qwen-3 0.6B             & 0.71 & 0.64 & 0.39 & 0.73 & 0.05 & 0.00 & 0.90 & 0.06 & 0.02 \\
Qwen-3 1.7B             & 0.62 & 0.64 & 0.31 & 0.76 & 0.25 & 0.16 & 0.91 & 0.16 & 0.12 \\
Qwen-3 4B               & 0.42 & 0.55 & 0.05 & 0.73 & 0.49 & 0.32 & 0.88 & 0.32 & 0.26 \\
Qwen-3 8B               & 0.57 & 0.61 & 0.24 & 0.70 & 0.54 & 0.33 & 0.79 & 0.29 & 0.20 \\
Qwen-3 14B              & 0.61 & 0.64 & 0.30 & 0.71 & 0.55 & 0.34 & 0.80 & 0.32 & 0.23 \\
Qwen-3 32B              & 0.73 & 0.71 & 0.48 & 0.71 & 0.55 & 0.35 & 0.80 & 0.33 & 0.25 \\
Gemma-3 4B              & 0.50 & 0.59 & 0.16 & 0.63 & 0.46 & 0.20 & 0.74 & 0.25 & 0.15 \\
Gemma-3 12B             & 0.61 & 0.64 & 0.31 & 0.75 & 0.04 & 0.03 & 0.92 & 0.11 & 0.09 \\
Gemma-3 27B             & 0.74 & 0.72 & 0.50 & 0.75 & 0.03 & 0.02 & 0.93 & 0.03 & 0.02 \\
LLaMA-3.2 1B            & 0.56 & 0.30 & 0.00 & 0.47 & 0.38 & 0.03 & 0.42 & 0.14 & 0.01 \\
LLaMA-3.2 3B            & 0.58 & 0.62 & 0.26 & 0.68 & 0.37 & 0.16 & 0.81 & 0.19 & 0.09 \\
LLaMA-3.1 8B            & 0.49 & 0.58 & 0.14 & 0.71 & 0.38 & 0.19 & 0.86 & 0.22 & 0.14 \\
LLaMA-3.1 70B           & 0.61 & 0.62 & 0.28 & 0.64 & 0.47 & 0.22 & 0.73 & 0.23 & 0.12 \\ \midrule
\textsc{saquet}         & ---    & ---    & ---    & 0.78  & 0.40  & 0.28  & \textbf{0.93}  & \textbf{0.52}  & \textbf{0.48}  \\ \midrule
Random (50/50)          & 0.50 & 0.42 & 0.00 & 0.50 & 0.34 & 0.00 & 0.50 & 0.13 & 0.00 \\
Always Not Flawed       & 0.64 & 0.00 & 0.00 & 0.74 & 0.00 & 0.00 & 0.93 & 0.00 & 0.00 \\
Always Flawed           & 0.37 & 0.54 & 0.00 & 0.26 & 0.41 & 0.00 & 0.08 & 0.14 & 0.00 \\ \bottomrule
\end{tabular}
\caption{\small Human--judge agreement for shortcuts and writing flaw detection. * on accuracy means the method has significantly better predictions than all trivial baselines \cite[Mcnemar's test, $p < 0.05$ with Bonferroni correction]{mcnemar1947note}.
Appendix~\ref{appendix:writing_flaw_results} has results grouped by each of the 19 writing flaws. The most reliable \mm{} judges are \textbf{bold}, informing \model{}'s design.}
\label{table:combined_shortcuts_writing}
\end{table*}

%% file: data/judge_contamination.tex
\begin{table}[]
\small
\centering
\renewcommand{\arraystretch}{0.7}
\setlength{\tabcolsep}{3.8pt}
\begin{tabular}{@{}lccc@{}}
\toprule
Method            & Accuracy & F1 Score & Cohen's $\kappa$ \\ \midrule
Google + Gemini Pro            & 0.70*   & 0.65   & 0.41 \\
Google + GPT-5           & \textbf{0.71}*   & \textbf{0.68}   & \textbf{0.44} \\
Google + Claude Sonnet            & 0.69*   & 0.64   & 0.40 \\ \midrule
Brave + GPT-5             & 0.54   & 0.28   & 0.14 \\
Perplexity + GPT-5        & 0.64   & 0.53   & 0.31 \\
Exa + GPT-5                & 0.59   & 0.43   & 0.22 \\
Tavily + GPT-5             & 0.58   & 0.41   & 0.20 \\
Serper + GPT-5             & 0.68   & 0.65   & 0.36 \\ \midrule
Random (50/50)    & 0.50   & 0.52   & 0.00 \\
Always Not Flawed & 0.46   & 0.00   & 0.00 \\
Always Flawed     & 0.54   & 0.70   & 0.00 \\ \bottomrule
\end{tabular}
\caption{\small Human--judge agreement for contamination detection. * means the method is significantly better than trivial baselines \cite[Mcnemar's test, $p < 0.05$, Bonferroni correction]{mcnemar1947note}.
Appendix~\ref{appendix:search_engine_results} shows all \mm{}/\textsc{api} combinations. Google with GPT-5 is the most reliable \mm{} judge.}
\label{table:judge_contamination}
\end{table}


%% file: figures/dataset_scores.tex
\begin{figure*}[t]
    \centering
    \includegraphics[width=\linewidth]{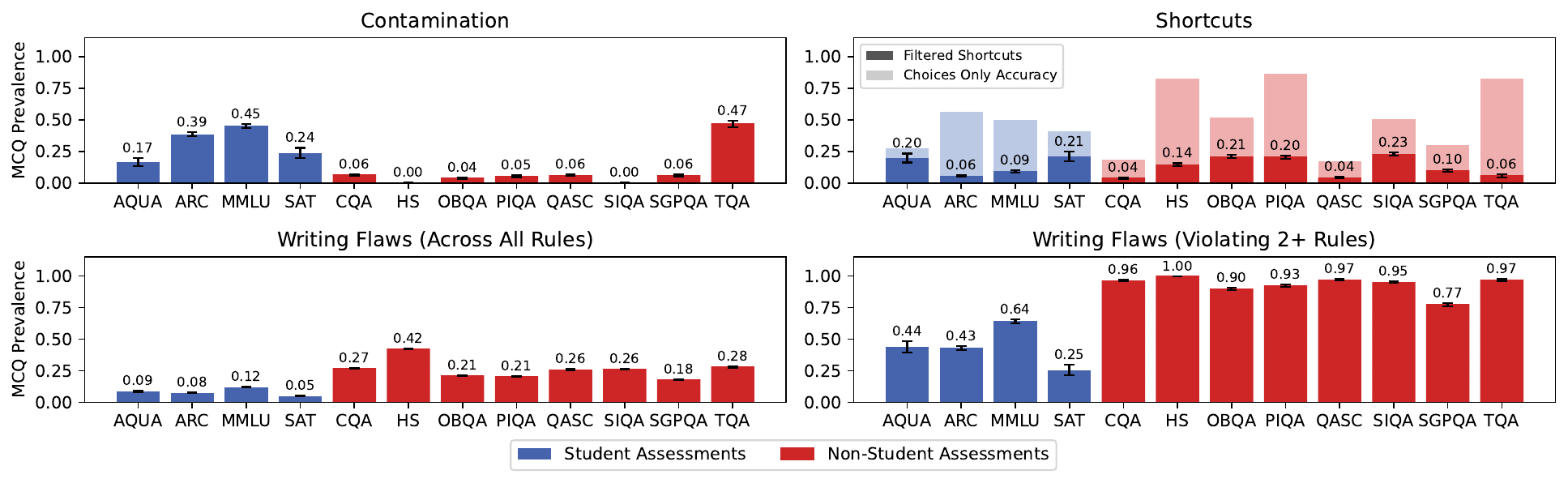}
    \caption{\label{fig:dataset_scores} \small Prevalence of flaws in \mcqa benchmarks, grouped by whether the \mcq{}s originate from student assessments.
    While \mcq{}s from exam-based benchmarks are more commonly found online (top left), they contain far fewer writing flaws (bottom).}
\end{figure*}

%% file: data/accuracy_impact.tex
\begin{table*}
\centering
\scriptsize
\renewcommand{\arraystretch}{1}
\setlength{\tabcolsep}{3pt}
\begin{tabular}{l|ccc|ccc|ccc|ccc}
\multicolumn{1}{c}{} & \multicolumn{3}{c}{\textit{Contamination}} & \multicolumn{3}{c}{\textit{Shortcuts}} & \multicolumn{3}{c}{\textit{2+ Writing Flaws}} & \multicolumn{3}{c}{\textit{Any Flaw}} \\
\toprule
Dataset & Flaw & No Flaw & $\Delta$Acc & Flaw & No Flaw & $\Delta$Acc & Flaw & No Flaw & $\Delta$Acc & Flaw & No Flaw & $\Delta$Acc \\
\midrule
AQUA & 0.74 $\pm$ 0.07 & 0.76 $\pm$ 0.02 & \positive{+2.89} & 0.75 $\pm$ 0.05 & 0.76 $\pm$ 0.02 & \positive{+0.65} & 0.68 $\pm$ 0.03 & 0.81 $\pm$ 0.02 & \positive{+19.34} & 0.72 $\pm$ 0.03 & 0.81 $\pm$ 0.02 & \positive{+11.52} \\
ARC & 0.90 $\pm$ 0.01 & 0.87 $\pm$ 0.01 & \negative{-3.31} & 0.85 $\pm$ 0.02 & 0.88 $\pm$ 0.00 & \positive{+4.03} & 0.86 $\pm$ 0.01 & 0.89 $\pm$ 0.01 & \positive{+3.62} & 0.88 $\pm$ 0.01 & 0.88 $\pm$ 0.01 & \positive{+0.08} \\
CQA & 0.74 $\pm$ 0.04 & 0.78 $\pm$ 0.01 & \positive{+4.71} & 0.70 $\pm$ 0.05 & 0.78 $\pm$ 0.01 & \positive{+11.49} & 0.77 $\pm$ 0.01 & 0.89 $\pm$ 0.03 & \positive{+14.60} & 0.77 $\pm$ 0.01 & 0.88 $\pm$ 0.04 & \positive{+14.12} \\
HS & 0.87 $\pm$ 0.03 & 0.78 $\pm$ 0.01 & \negative{-9.84} & 0.74 $\pm$ 0.02 & 0.79 $\pm$ 0.01 & \positive{+6.75} & 0.78 $\pm$ 0.01 & --- & --- & 0.78 $\pm$ 0.01 & --- & --- \\
MMLU & 0.85 $\pm$ 0.01 & 0.75 $\pm$ 0.01 & \negative{-11.20} & 0.76 $\pm$ 0.03 & 0.80 $\pm$ 0.01 & \positive{+4.95} & 0.77 $\pm$ 0.01 & 0.83 $\pm$ 0.01 & \positive{+7.83} & 0.79 $\pm$ 0.01 & 0.79 $\pm$ 0.02 & \negative{-0.83} \\
OBQA & 0.86 $\pm$ 0.04 & 0.87 $\pm$ 0.01 & \positive{+1.69} & 0.86 $\pm$ 0.01 & 0.87 $\pm$ 0.01 & \positive{+1.45} & 0.86 $\pm$ 0.01 & 0.92 $\pm$ 0.01 & \positive{+6.33} & 0.87 $\pm$ 0.01 & 0.92 $\pm$ 0.01 & \positive{+6.71} \\
PIQA & 0.90 $\pm$ 0.03 & 0.91 $\pm$ 0.01 & \positive{+0.73} & 0.86 $\pm$ 0.01 & 0.92 $\pm$ 0.01 & \positive{+6.16} & 0.90 $\pm$ 0.01 & 0.93 $\pm$ 0.01 & \positive{+3.40} & 0.90 $\pm$ 0.01 & 0.94 $\pm$ 0.01 & \positive{+3.95} \\
QASC & 0.59 $\pm$ 0.05 & 0.60 $\pm$ 0.01 & \positive{+2.96} & 0.55 $\pm$ 0.05 & 0.61 $\pm$ 0.01 & \positive{+9.33} & 0.60 $\pm$ 0.01 & 0.70 $\pm$ 0.06 & \positive{+17.20} & 0.60 $\pm$ 0.01 & 0.66 $\pm$ 0.07 & \positive{+9.00} \\
SAT & 0.74 $\pm$ 0.04 & 0.79 $\pm$ 0.01 & \positive{+6.65} & 0.77 $\pm$ 0.04 & 0.78 $\pm$ 0.02 & \positive{+0.99} & 0.74 $\pm$ 0.05 & 0.79 $\pm$ 0.01 & \positive{+6.71} & 0.77 $\pm$ 0.03 & 0.80 $\pm$ 0.01 & \positive{+3.91} \\
SIQA & 1.00 $\pm$ 0.00 & 0.80 $\pm$ 0.01 & \negative{-19.86} & 0.80 $\pm$ 0.02 & 0.80 $\pm$ 0.01 & \positive{+0.12} & 0.79 $\pm$ 0.01 & 0.97 $\pm$ 0.01 & \positive{+22.42} & 0.79 $\pm$ 0.01 & 0.97 $\pm$ 0.01 & \positive{+22.74} \\
SGPQA & 0.47 $\pm$ 0.03 & 0.48 $\pm$ 0.01 & \positive{+2.49} & 0.53 $\pm$ 0.02 & 0.47 $\pm$ 0.01 & \negative{-9.71} & 0.45 $\pm$ 0.01 & 0.57 $\pm$ 0.01 & \positive{+25.85} & 0.46 $\pm$ 0.01 & 0.57 $\pm$ 0.01 & \positive{+24.02} \\
TQA & 0.76 $\pm$ 0.02 & 0.78 $\pm$ 0.02 & \positive{+2.47} & 0.60 $\pm$ 0.06 & 0.78 $\pm$ 0.01 & \positive{+31.52} & 0.77 $\pm$ 0.01 & 0.93 $\pm$ 0.03 & \positive{+21.53} & 0.77 $\pm$ 0.01 & 0.94 $\pm$ 0.04 & \positive{+21.95} \\
\midrule
\textbf{Micro $\mu$} & 0.81 $\pm$ 0.01 & 0.76 $\pm$ 0.00 & \negative{-6.90} & 0.77 $\pm$ 0.01 & 0.77 $\pm$ 0.00 & \negative{-1.03} & 0.75 $\pm$ 0.00 & 0.83 $\pm$ 0.00 & \positive{+9.83} & 0.76 $\pm$ 0.00 & 0.80 $\pm$ 0.01 & \positive{+5.39} \\
\textbf{Macro $\mu$} & 0.78 $\pm$ 0.04 & 0.76 $\pm$ 0.03 & \negative{-2.48} & 0.73 $\pm$ 0.03 & 0.77 $\pm$ 0.04 & \positive{+5.35} & 0.75 $\pm$ 0.04 & 0.84 $\pm$ 0.04 & \positive{+12.17} & 0.76 $\pm$ 0.04 & 0.83 $\pm$ 0.04 & \positive{+9.61} \\
\bottomrule
\end{tabular}
\caption{\small \mm{} accuracy on flawed and not flawed \mcqa{} dataset splits across each flaw type. Micro/macro averages show that contaminated splits tend to have higher accuracy, and splits with two or more writing errors tend to have lower accuracy. \label{table:accuracy_flaw}}
\end{table*}

%% file: 2026_arr_metabench/sections/40_results.tex
\input{data/ranking_imapct}

\section{A Report Card for \mcqa{} Benchmarks} \label{section:results}

Having validated \model{}, we now use it to audit benchmarks (\cref{subsection:eval_benchmark}) and study how flaws impact evaluation (\cref{subsection:impact}, \cref{subsection:rankings}).
We reveal writing errors are rife (\cref{subsection:which_writing_flaws}), unaffected by prior fixes (\cref{subsection:fixes}).

\subsection{\mcqa{} Benchmarks are Rife with Flaws} \label{subsection:eval_benchmark}

We run \model{} on up to 1000 sampled \mcq{}s from the test sets of 12 benchmarks in Table~\ref{table:datasets}, predicting contamination, shortcut, and writing errors.
All datasets have flaws (Figure~\ref{fig:dataset_scores}): we detect 47\% of TruthfulQA exists online, 23\% of SocialIQA has shortcuts, and on average, each item in HellaSwag violates 44\% of the 19 writing rules.
In education, \mcq{}s violating 2+ writing rules are ``unacceptable'' \citep{tarrant2006frequency}, but we suspect 7/12 datasets have over 90\% of items with 2+ writing violations.

Grouping items by their origin shows those from educator-written student exams (\textcolor{blue}{blue}, not hatched) have fewer writing flaws than those written automatically or by crowdworkers (\textcolor{red}{red}, hatched), showing education's value in \mcqa{} design.
Out of the non-educator \mcq{}s, SGPQA has the fewest writing flaws; experts wrote them with \mm{}s, so human-\textsc{ai} writing is a promising path to improve \mcq{}s.
Despite this benefit, educator-written \mcq{}s are more contaminated; annotation (\cref{subsection:annotations}) found many items online as study aids (e.g., flashcards, tutor sites),~so \mm{} developers could filter websites linked~to test preparation to stop this \cite{soldaini2024dolma}.
One may expect contamination to link to release dates, but older \mcq{}s (HellaSwag) have 0~contamination score while more recent ones (TQA) can be higher. 

Lastly, choices-only accuracy can overestimate shortcut prevalence (Fig~\ref{fig:dataset_scores}, top right).~While~ARC, TQA, and OBQA have high choices-only accuracy, it often stems from non-problematic strategies---the inferred question often matches the original.
After filtering these cases, shortcut prevalence drops (e.g., $83\% \rightarrow 6\%$ on TQA).
Past work cites high~choices-only accuracy as evidence of dataset flaws \citep{chandak2025answer}, but without considering \textit{why} models succeed, this metric alone overestimates them.

\subsection{\mcqa{} Flaws Degrade \mm{} Evaluation} \label{subsection:impact}

Having exposed benchmark flaws, we now test~their evaluation impact, informing which issues~to prioritize fixing.
We evaluate 10 \mm{}s---GPT, Gemini, Claude, and Cohere \cref{subsection:baselines}---on our \mcq{}s.~Given the brittleness of \mm{}s' first-token probabilities~\citep{wang2024look, wang2024my, molfese-etal-2025-right}, we instead use a prompt that requests a structured response with the model's selected choice, implemented in InspectAI (Appendix~\ref{appendix:inspect}).
We report mean \mm{} accuracy \footnote{Mean accuracy is used in \nlp{} \cite{hofmann2025fluid}, but using difficulty from Item Response Theory \cite{lord2008statistical}, an education tool, maintains claims (Appendix~\ref{appendix:irt_results}).} on the ``Flaw'' vs ``No Flaw'' splits~(\cref{subsection:eval_benchmark}) to analyze how each flaw type relates to accuracy.

Scores differ per-dataset (Table~\ref{table:accuracy_flaw}), but: \textbf{1)~Contaminated splits have higher accuracy}, similar to how~memorized items~are often easier~for students and \mm{}s \cite{ebbinghaus_memory_1913, sainz-etal-2023-nlp}; \textbf{(2) Splits violating 2+ writing rules have~lower accuracy}, aligning with research revealing poorly written items mislead test-takers \citep{schmucker2025impact, nahum-etal-2025-llms}; and while \citet{gupta2024improving} find filtering \mcq{}s with choices-only success lowers \mcqa{} accuracy,\footnote{We reproduce this result in Appendix~\ref{appendix:shortcut_ablation}.} they do not consider \textit{how} success arises.
But \textbf{(3) Splits with shortcuts have similar/mixed accuracy after correcting for strategy}, backing claims that choices-only success alone overstates shortcut issues \citep{balepur2025test}.
Viewing all flaws, writing errors~tend to dominate (Table~\ref{table:accuracy_flaw}, right), so lower \mcqa{} scores could stem from model failures in solving poorly-written items, rather than what \mcqa{} aims to test.

Across every flaw, we argue that writing errors are most critical to fix, given their accuracy drops (Table~\ref{table:accuracy_flaw}) and prevalence (Fig~\ref{fig:dataset_scores}).
We study~the~most common writing flaws in \cref{subsection:which_writing_flaws} to guide future~work.

\subsection{\mcqa{} Flaws Can Shift \mm{} Rankings} \label{subsection:rankings}

Users look at \mcqa{} benchmark rankings to decide which \mm{}s to use daily \cite{liang2023holistic}, and researchers use them to select model checkpoints for further training \cite{olmo20242}. 
To study~if \mcq{} flaws could change these decisions, we run the 10 \mm{}s in \cref{subsection:impact} over all 12 benchmarks using the ``Full'' and ``No Flaw'' data splits.
We also make a ``Random'' split by uniformly sampling the same number of \mcq{}s as in the ``No Flaw'' split,\footnote{Averaged across 100 random seeds (0--99) for robustness.} testing whether changes are just due to sampling variation.

\mm{} ranks on the ``Full'' and ``No Flaw'' splits are identical for contamination and shortcuts (Table~\ref{table:ranking_change}), but ranks shift by up to two positions for writing errors.
These shifts exceed random sampling: writing errors can change the models users and researchers select. We speculate these changes are starker when comparing \mm{}s of similar abilities (e.g., \mm{} pre-training checkpoints).
This further backs our suggestion for \nlp{} to devote more time toward reducing writing errors in benchmarks.

\subsection{The Writing Errors for \nlp{} to Address} \label{subsection:which_writing_flaws}


Writing errors are pervasive (\cref{subsection:eval_benchmark}) and shift \mm{} ranks (\cref{subsection:impact}), so we advise researchers to focus on fixing them.
To inform these efforts, we now study the most common writing errors in our datasets.
On student-based (standardized tests) and non-student-based \mcq{}s, five flaws emerge: ambiguous question stems, indirectly asking questions, unclear language, grammatical inconsistency in questions and choices, and implausible distractors (Figure~\ref{fig:writing_flaws}).
The first four lead to unclear \mcq{}s, an issue text simplification research has been working to remedy \cite{chandrasekar-etal-1996-motivations}.
The last links~to distractor difficulty, matching recent education efforts in \mm{} distractor generation \cite{mcnichols2023automated, lee-etal-2025-generating-plausible}.
As similar writing errors plague \mcq{}s in \nlp{} and education---and both are building solutions---there is a clear opportunity for the fields to collaborate on~tools to fix these flaws.

\input{figures/writing_flaws}
\input{data/fixes}

\subsection{The Trade-Offs of Benchmark Fixes} \label{subsection:fixes}

Prior work also finds writing errors in \mcqa{} benchmarks, spurring new versions.
MMLU-Pro \cite{wang2024mmlu} and MMLU-Redux \cite{gema2024we} adjust MMLU to fix ambiguity and gold answers; GoldenSwag revises HellaSwag to improve grammar and distractors \cite{chizhov2025hellaswagvaliditycommonsensereasoning}.
We now run \model{} to study these revisions.

We first verify each revision works: GoldenSwag and MMLU-Redux fix labels, boosting accuracy, while MMLU-Pro drops accuracy via \mm{}-written distractors (Table~\ref{table:dataset_fixes}).
But GoldenSwag and MMLU-Pro add writing errors, which can impact evaluation (\cref{subsection:impact}).
Analyzing writing rules helps explain~why.
GoldenSwag removes \mcq{}s with multiple correct answers ($18\% \rightarrow 4\%$) and differences in choice length ($52\% \rightarrow 43\%$) as intended, but adds~grammar inconsistencies ($68\% \rightarrow 79\%$), perhaps due to its automated filtering.
MMLU-Pro's \mm{}-written distractors lower accuracy, but we detect they are less plausible ($7\% \rightarrow 17\%$) and correct when they should be incorrect ($10\% \rightarrow 22\%$).
For example, one MMLU-Pro \mcq{} asks for an element's change in atomic number after emitting particles, with answer ``zero''.  
\mm{}s incorrectly create the distractor ``does not change''; this lowers accuracy, but only because models are split on which answer to select.

Finally, we test whether \mm{}s can rewrite \mcq{}s to reduce the writing errors \model{} detects.
We prompt GPT-5.2, Claude Sonnet, and Qwen-235B with \mcq{}s from TruthfulQA and the writing errors detected by \model{}, and ask each model to return an \mcq{} that corrects these flaws.
Across \mm{}s, this does not eliminate errors and can even introduce new ones (Table~\ref{appendix:table:rewrite}), motivating the need for future work to explore novel approaches beyond simple prompting for rewriting \mcq{}s.



While these revisions take meaningful steps toward reliable \mcqa{} evaluation, they highlight that benchmark correction is multi-objective: fixing one flaw can add others.
Robust \mcqa{} correction thus requires iterative refinement, and \model{} is one way to track progress. Researchers can use our tool to verify whether their targeted fixes improve \mcq{}s as intended or inadvertently add new errors.

%% file: data/ranking_imapct.tex
\begin{table*}
\centering
\scriptsize
\renewcommand{\arraystretch}{1}
\setlength{\tabcolsep}{2pt}

\begin{tabular}{l|ccc|ccc|ccc|ccc}
\multicolumn{1}{c}{} & \multicolumn{3}{c}{\textit{Contamination}} & \multicolumn{3}{c}{\textit{Shortcuts}} & \multicolumn{3}{c}{\textit{2+ Writing Errors}} & \multicolumn{3}{c}{\textit{Any Flaw}} \\
\toprule
Model & All & No Flaw & Random & All & No Flaw & Random & All & No Flaw & Random & All & No Flaw & Random \\
\midrule
Gemini-2.5 Lite & 0.530 (10) & 0.525 (10) & 0.530 (10) & 0.530 (10) & 0.525 (10) & 0.520 (10) & 0.530 (10) & 0.542 (10) & 0.489 (10) & 0.530 (10) & 0.505 (10) & 0.469 (10) \\
Gemini-2.5 Flash & 0.562 (9) & 0.560 (9) & 0.568 (9) & 0.562 (9) & 0.559 (9) & 0.554 (9) & 0.562 (9) & 0.585 (9) & 0.526 (9) & 0.562 (9) & 0.557 (9) & 0.514 (9) \\
Gemini-2.5 Pro & 0.865 (3) & 0.857 (3) & 0.856 (3) & 0.865 (3) & 0.862 (3) & 0.862 (3) & 0.865 (3) & 0.961 (1) & 0.909 (2) & 0.865 (3) & 0.955 (1) & 0.894 (2) \\
GPT-5 Nano & 0.825 (6) & 0.815 (6) & 0.816 (6) & 0.825 (6) & 0.823 (6) & 0.822 (6) & 0.825 (6) & 0.927 (5) & 0.868 (6) & 0.825 (6) & 0.911 (5) & 0.850 (6) \\
GPT-5 Mini & 0.858 (4) & 0.848 (4) & 0.849 (4) & 0.858 (4) & 0.858 (4) & 0.855 (4) & 0.858 (4) & 0.953 (3) & 0.899 (4) & 0.858 (4) & 0.942 (3) & 0.883 (4) \\
GPT-5 & 0.878 (1) & 0.870 (1) & 0.871 (1) & 0.878 (1) & 0.877 (1) & 0.875 (1) & 0.878 (1) & 0.958 (2) & 0.916 (1) & 0.878 (1) & 0.950 (2) & 0.902 (1) \\
Claude 4.5 Haiku & 0.832 (5) & 0.821 (5) & 0.823 (5) & 0.832 (5) & 0.829 (5) & 0.828 (5) & 0.832 (5) & 0.919 (6) & 0.872 (5) & 0.832 (5) & 0.898 (6) & 0.852 (5) \\
Claude 4.5 Sonnet & 0.874 (2) & 0.863 (2) & 0.864 (2) & 0.874 (2) & 0.872 (2) & 0.871 (2) & 0.874 (2) & 0.937 (4) & 0.908 (3) & 0.874 (2) & 0.916 (4) & 0.890 (3) \\
Command R & 0.699 (8) & 0.686 (8) & 0.694 (8) & 0.699 (8) & 0.699 (8) & 0.694 (8) & 0.699 (8) & 0.730 (8) & 0.705 (8) & 0.699 (8) & 0.676 (8) & 0.678 (8) \\
Command R+ & 0.720 (7) & 0.706 (7) & 0.712 (7) & 0.720 (7) & 0.719 (7) & 0.716 (7) & 0.720 (7) & 0.766 (7) & 0.738 (7) & 0.720 (7) & 0.719 (7) & 0.708 (7) \\
\midrule
\textbf{Spearman's $\rho$} & --- & 1.000 & 1.000 & --- & 1.000 & 1.000 &  & 0.927 & 0.986 & --- & 0.927 & 0.978 \\
\bottomrule
\end{tabular}

\caption{\small \mm{} rank (in parentheses) correlation between full vs no flaw/random splits. Removing contamination and shortcuts does not shift rankings, but filtering writing errors or any flaw does beyond random, confirmed via permutation tests ($\alpha = 0.01$). \label{table:ranking_change}}
\end{table*}

%% file: figures/writing_flaws.tex
\begin{figure}[t]
    \centering
    \includegraphics[width=\linewidth]{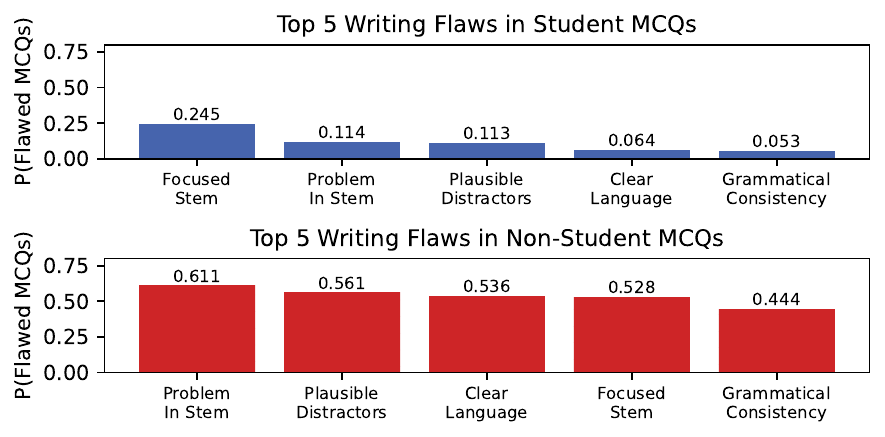}
    \caption{\label{fig:writing_flaws} \small The five most common writing errors \model{} predicts in \mcqa{} benchmarks, grouped by whether they stem from student exams. Most flaws relate to clarity and distractor difficulty. Appendix~\ref{appendix:writing_flaw_results} has the full distribution of 19 flaws.}
\end{figure}

%% file: data/fixes.tex
\begin{table}[]
\small
\centering
\setlength{\tabcolsep}{3.5pt}
\begin{tabular}{@{}lcccc@{}}
\toprule
\textbf{Dataset} &
  \multicolumn{1}{l}{\textbf{Contam.}} &
  \multicolumn{1}{l}{\textbf{Shortcuts}} &
  \multicolumn{1}{l}{\textbf{Writing}} &
  \multicolumn{1}{l}{\textbf{Accuracy}} \\ \midrule
HellaSwag  & 0.00 & 0.31 & 0.42 & 0.78 \\
GoldenSwag & 0.00 & 0.34 & 0.44 & 0.84 \\ \midrule
MMLU       & 0.45 & 0.12 & 0.12 & 0.79 \\
MMLU Pro   & 0.24 & 0.12 & 0.15 & 0.63 \\
MMLU Redux & 0.44 & 0.13 & 0.12 & 0.81 \\ \bottomrule
\end{tabular}
\caption{\small \model{} scores on \mcqa{} datasets and their revised versions, along with the average accuracy of the 10 \mm{}s in \cref{subsection:impact}. We score writing errors over all 19 rules. Revised datasets fix what they intend to (e.g. lower MMLU-Pro accuracy), but add errors (e.g. writing flaws in MMLU-Pro). \label{table:dataset_fixes}}
\end{table}

\begin{table}[h]
\small
\centering
\setlength{\tabcolsep}{3.5pt}
\begin{tabular}{lccc}
\toprule
Model & \probP(2+ Errors) & \probP(Flawless) & \probP(New Error) \\
\midrule
GPT-5.2        & 0.97 $\rightarrow$ 0.64 & 0.26 & 0.56 \\
Claude Sonnet  & 0.97 $\rightarrow$ 0.64 & 0.32 & 0.68 \\
Qwen 235B      & 0.97 $\rightarrow$ 0.68 & 0.26 & 0.62 \\
\bottomrule
\end{tabular}
\caption{\small The proportion of \mcq{}s with 2+ writing errors, no writing errors, and new errors after prompting GPT, Claude, and Qwen to fix writing errors in TruthfulQA. Simple prompting does not fully reduce errors and often introduces new ones.}
\label{appendix:table:rewrite}
\end{table}

%% file: 2026_arr_metabench/sections/50_related_work.tex
\section{Related Work} \label{section:related_work}

As we instantiate \mcqa{} educational theory via \nlp{} methods, we review \mcqa{}'s history in both \nlp{} evaluation (\cref{subsection:mcqa_nlp}) and human assessments (\cref{subsection:mcqa_human}) 

\subsection{Multiple-Choice Evaluation in \nlp{}} \label{subsection:mcqa_nlp}

Multiple-choice questions (\mcq{}s) have historically been used in \nlp, with early work testing commonsense \cite{levesque2012winograd} and comprehension \cite{richardson2013mctest}; solving these was an ``\textsc{ai} grand challenge'' \cite{reddy1988foundations}.
\mcqa{} became standard with the advent of \mm{}s; \citet{robinson2023leveraging} found one could prompt~\mm{}s to answer \mcq{}s like students and easily score them, spurring harder \mcq{}s \cite{rein2023gpqa} and leaderboard use \cite{liang2023holistic, open-llm-leaderboard-v2}.

Despite this popularity, many \nlp{} works~show issues in \mcqa{}.
Models unreliably solve \mcq{}s---brittle to symbolic \cite{alzahrani2024benchmarks}, logical \cite{kawabata2023evaluating, balepur-etal-2024-easy, balepur-etal-2025-reverse}, and language \cite{singh2024global} perturbations.
\mcqa{} datasets fall victim to contamination \cite{li-etal-2024-open-source},~poor grammar \cite{mousavi2025garbage}, plausibility errors \cite{Palta2024PlausiblyPQ}, and shortcuts \cite{Balepur2024ArtifactsOA, balepur-rudinger-2024-large}.
Prior research also argues that \mcqa{} misaligns with educational assessment goals like commonsense \cite{davis2014limitations}, neglect real user needs \cite{saxon2024benchmarks, balepur-etal-2025-best}, and should be categorized by what the question intends to evaluate \citep{rodriguez-boyd-graber-2021-evaluation, rogers2023qa}.
We synthesize these insights to prioritize flaws in \model{}'s design.

In response, previous work has designed tools to improve \mcqa{} benchmarks globally in diversity \cite{perlitz2024these}, efficiency \cite{hofmann2025fluid}, and saturation \cite{polo2024tinybenchmarks}, and at the \mcq{} level via dataset-specific annotation protocols \cite{wang2024mmlu, chizhov2025hellaswagvaliditycommonsensereasoning, mousavi2025garbage}.
Conversely, we present \model{} as a general toolkit for the latter and use it to audit 12 \nlp{} benchmarks, study how flaws impact \mm{} evaluation, and analyze prior correction strategies.

\subsection{Multiple-Choice Testing in Education} \label{subsection:mcqa_human}

\mcqa{} is a long-standing format for students \cite{monroe1917report}, but education researchers still~look~to boost its construct validity \cite{cronbach1955construct}: ensuring \mcqa{} tests what it intends to.
This has been achieved via new scoring \cite{finetti1965methods}, answer formats \cite{snow2012construct}, and adaptive testing \cite{lord1964effect}---all validated with students.
We argue \nlp{} can mirror~this---testing how educational theory alters evaluation---as done in \model{}.

The closest work to ours is \mcq{} quality estimation \cite{wang2023automatic}.
Such work often relies on similarity metrics like \textsc{bleu} \cite{mulla2023automatic}, custom rules \cite{moore2023assessing}, or~item measures like perplexity \cite{raina2022multiple}, but these disagree with expert judgments \cite{van2021human}.
Recent work has thus extended stronger \nlp{} methods like \mm{}-as-a-judge \cite{moore2024automatic}, but usually target a single flaw type.
In contrast, \model{} flags contamination, shortcuts, and writing errors, and we release all code and annotations to aid future work in this sparse~area.

%% file: 2026_arr_metabench/sections/60_conclusion.tex
\section{Conclusion: \model{}'s Next Steps} \label{section:conclusion}

Multiple-choice benchmarks are laced with flaws that harm \nlp{}, but \model{} offers a path~towards redemption: predicting \mcq{}s with~contamination, shortcuts, and writing errors.
While \model{}'s \mm{} judges are sufficient for benchmark audits, further work remains in executing contamination detection with cheaper search \textsc{api}s, training efficient \mm{}s to match closed-source judges, and stress-testing \model{}'s generalization across languages and domains.
Our released code and annotated validation sets will facilitate these efforts.

Beyond diagnosis, our future work seeks to repair these flaws; we believe optimizing prompts for rewriting via \model{}-as-a-verifier \cite{opsahl-ong-etal-2024-optimizing}, running user studies in human–\textsc{ai} rewriting interfaces \cite{cui2024promises, wallace-etal-2019-trick, sung-etal-2025-benchmark}, and drawing on educational testing theory for construct validation \citep{rodriguez-etal-2021-evaluation, hofmann2025fluid} are useful next steps towards this.
Overall, despite \nlp{} facing an ``evaluation crisis'' \cite{liao2023rethinking}, our paper shows educational standards are a valuable lifeline for rigorously assessing \nlp{} systems.

%% file: 2026_arr_metabench/sections/70_limitation_ethics.tex
\section{Limitations} \label{section:limitations}

\model{} is currently designed to detect flaws in \nlp{} benchmarks, but does not currently offer remediation strategies apart from flagging the item for human review or completely discarding the item.
While still useful for improving \nlp{} benchmarks, we believe a necessary future step is using these scores to refine \mcq{}s.
This step is outside the scope of our paper, but there is extensive research in automatically generating \mcq{}s with \mm{}s \cite{lee-etal-2025-generating-plausible, parikh-etal-2025-lookalike}, and we are excited about applying these insights to future iterations of \model{}, incorporating \mm{} rewrites for these flaws without solely relying on human intervention.

While we have tested \model{}'s generalization across our 12 \mcqa{} benchmarks spanning different domains, difficulties, and creation strategies (Table~\ref{table:datasets}), there are other types of \mcqa{} datasets we did not explore, such as languages beyond English \cite{son2024kmmlu, li2023cmmlu}, specific domains like medicine \cite{pal2022medmcqa} and coding \cite{gu2024cruxeval}, and cultures \cite{chiu-etal-2025-culturalbench}.
Some dimensions may arise in multi-domain benchmarks like MMLU \cite{hendrycks2020measuring} and Super GPQA \cite{du2025supergpqa}, but they were not a central focus of our experiments.
By releasing~our toolkit publicly, we hope to collect feedback from the \nlp{} community on which areas \model{} struggles in and learn how it can be improved.

There are are many other issues in \mcqa{} benchmarks that \model{} does not tackle, particularly at the global level like saturation \cite{saxon2024benchmarks}, efficiency \cite{perlitz2023efficient}, and diversity \cite{singh2024global}.
To narrow \model{}'s scope, we draw on prior critiques of \mcqa{} evaluations \cite{balepur-etal-2025-best, chandak2025answer} and education research \cite{haladyna1989taxonomy} and focus on item-specific errors, leading to our flaws of contamination, shortcuts, and writing errors.
As \model{} uses the InspectAI library \cite{inspect_ai_framework}, it is simple for researchers to extend our tool and add metrics they value, designed as ``scorers'' in the library.

Finally, some argue that benchmark errors (e.g. poor grammar) are representative of real-world user queries, so they do not need to be remedied.
We counter that if reasoning over noisy inputs is part of the task description, these should still be introduced systematically and researchers should know which items have these flaws---ideally as a separate task \cite{guo-vosoughi-2024-disordered}---so researchers can better understand where their models fail.
\model{} facilitates these evaluation efforts by detecting such flaws, allowing researchers to create clean and noisy splits of their benchmarks (\cref{subsection:impact}).

\section{Ethical Considerations}

Low-quality benchmarks can undermine \nlp{} evaluations, and \model{} takes steps to correct that in \mcqa{} datasets, offering a toolkit to flag flaws in \mcq{}s.
However, our \mm{} judges are imperfect (\cref{section:setup}), so~we advise against using \model{} as the only tool to flag and fix quality errors of \mcqa{} benchmarks, especially without any human intervention.
As discussed in \cref{section:conclusion}, we are excited about integrating \model{} into online user interfaces and running studies to understand how our toolkit can best support \nlp{} researchers and educators.

Generative AI (GenAI) was used in this project.
We used Cursor\footnote{\url{https://cursor.com/agents}} to design plots and refactor code, and GPT-5 to refine paper writing for brevity.
We never use GenAI for writing text from scratch in this paper.
We take complete responsibility for any GenAI errors.
By discussing GenAI usage here, we aim to encourage other researchers to do the same.

\section*{Acknowledgments}

We would like to thank the \abr{clip} lab at the University of Maryland for their support.
In particular, we thank Ayush Jhaveri, Paiheng Xu, Alexander Hoyle for reviews and discussions on earlier versions of this paper draft.
This material is based upon work supported by the National Science Foundation under \abr{iis}-2339746 (Rudinger) \abr{iis}-2403436 (Boyd-Graber), and \abr{dge}-2236417 (Balepur).
Any opinions, findings, and conclusions or recommendations expressed in this material are those of the author(s) and do not necessarily reflect the views of the National Science Foundation.
Access to Cohere models (Command-R, Command-R Plus)~was made possible via a Cohere for AI Research Grant.

%% file: 2026_arr_metabench/sections/100_appendix.tex
\section{Appendix} \label{section:appendix}

\subsection{Survey of \nlp{} Evaluation} \label{appendix:survey}

To motivate our design of \model{}, we first survey \textsc{ai} papers that release \mcqa{} benchmarks.
We search Google Scholar, Semantic Scholar, the ACL Anthology, and query Deep Research tools (ScholarQA) with keywords related to ``multiple-choice'' and ``benchmarks'', yielding 39 total papers from 2013--2025 for manual review.
For each paper, we mark: 1) whether the authors report any dataset quality control; and 2) whether the authors contextualize this quality with respect to other datasets.

We find that 23\% of benchmark papers report no quality control, and 49\% do not compare dataset quality to other datasets.
While comparing dataset quality is nice to have, we believe reporting quality control is necessary---fortunately, \model{} aids both of these goals.
Several works that draw directly from human exams report no quality control, assuming they are high-quality, but our analysis reveals that even these questions can have flaws (\cref{subsection:eval_benchmark}).
Thus, we recommend that all researchers review their \mcqa{} benchmarks before release.

\subsection{InspectAI Implementation} \label{appendix:inspect}

InspectAI\footnote{https://inspect.aisi.org.uk/} is a recent effort from the United Kingdom's AI Security Institute to standardize \nlp{} evaluations \cite{inspect_ai_framework}.
Any InspectAI framework contains three parts:
\begin{enumerate}[nosep]
    \item \textbf{Task:} The data for the task. In our setup, these are the question stem, choices, and answer of the multiple-choice question.
    \item \textbf{Solver:} The \nlp{} system that solves the task. In our setup, this is either a function that returns the entire dataset when we are scoring the dataset itself, or these are the \mm{}s from \cref{subsection:impact} that we run \mm{}s on our dataset.
    \item \textbf{Scorer:} How task success/failure is evaluated. In our setup, these are either the contamination, shortcuts, and writing flaw judges used in \model{}, or a standard accuracy score when we run \mm{}s on our dataset.
\end{enumerate}

InspectAI allows researchers to easily add their own tasks, solvers, and scorers in a standardized way, which in our case, will allow researchers to extend \model{} more easily.
It also provides an easy-to-use \textsc{ui}, which could form the basis for future user studies in \mm{} evaluation (Figure~\ref{fig:appendix:inspect}).
The library has been adopted in other benchmarking efforts like AstaBench \cite{bragg2025astabench}, and we use it in \model{} to motivate further use.
Appendix~\ref{appendix:prompts} has the prompts used in Inspect.

\subsection{Dataset Details} \label{appendix:dataset}

Out of the datasets in the paper, we use the training sets of each dataset for judge validation, and the test sets of each dataset for auditing.
The commonsense datasets (e.g., SocialIQA) have fully held-out test sets, so we instead use the validation sets; we find it to be a positive sign that the validation sets have low contamination rates, as this is likely an upper bound for test set contamination.
For datasets with more than 1000 \mcq{}s, we sample 1000 uniformly for auditing.
Exceptions are AQuA, SAT, and TruthfulQA, which only have a single set of questions, so we split them evenly between train and test, yielding 127 test set items for AQuA, 104 test set items for SAT, and 409 test set items for TruthfulQA.
We take random samples, so our metrics are unbiased estimates of the full data.

All datasets are publicly available, so our experiments are within their intended use.
We did not collect any datasets, so we did not check for PII.
To our knowledge, all questions are in English.

\subsection{\mm{} Implementation Details} \label{appendix:implementation}

We run every \mm{}s using default parameters, both for judge experiments and benchmark audits.
We use CPUs only when running \mm{}s via \textsc{api}s, one \textsc{nvidia} rtxa6000 \textsc{gpu} for open-weight \mm{}s below 8B parameters called via Huggingface, and eight \textsc{nvidia} rtxa5000's for all other open-weight \mm{}s called via Huggingface.
We allocate 24 hours for each run.
All results are reported from a single run.

In our experiments, we run all closed-source and Cohere models using litellm\footnote{https://www.litellm.ai/} and all other models via Huggingface's inference endpoint.\footnote{https://huggingface.co/docs/inference-endpoints/en/index} The \textsc{api}s and Huggingface endpoints for our \mm{}s are:
\begin{itemize}[nosep]
\item openai/gpt-5-nano-2025-08-07
\item openai/gpt-5-mini-2025-08-07
\item openai/gpt-5-2025-08-07
\item anthropic/claude-haiku-4-5-20251001
\item anthropic/claude-sonnet-4-5-20250929
\item gemini/gemini-2.5-pro
\item gemini/gemini-2.5-flash-lite
\item gemini/gemini-2.5-flash
\item cohere/command-r-08-2024
\item cohere/command-r-plus-08-2024
\item Qwen/Qwen3-0.6B
\item Qwen/Qwen3-1.7B
\item Qwen/Qwen3-4B
\item Qwen/Qwen3-8B
\item Qwen/Qwen3-14B
\item Qwen/Qwen3-32B
\item google/gemma-3-4b-it
\item google/gemma-3-12b-it
\item google/gemma-3-27b-it
\end{itemize}

\noindent The \mm{} used to evaluate each of the 19 writing flaws are as follows:

\begin{itemize}[nosep]
\item avoid\_k\_type: google/gemini-2.5-flash-lite,
\item avoid\_negatives: anthropic/claude-sonnet-4-5-20250929
\item avoid\_repetition: google/gemini-2.5-pro
\item clear\_language: openai/gpt-5-2025-08-07
\item equal\_length\_options: google/gemini-2.5-pro
\item focused\_stem: google/gemini-2.5-pro
\item grammatical\_consistency: google/gemini-2.5-pro
\item no\_absolute\_terms: google/gemini-2.5-pro
\item no\_all\_of\_the\_above: anthropic/claude-sonnet-4-5-20250929
\item no\_convergence\_cues: anthropic/claude-sonnet-4-5-20250929
\item no\_extraneous\_info: openai/gpt-5-2025-08-07
\item no\_fill\_in\_blank: google/gemini-2.5-pro
\item no\_logical\_cues: openai/gpt-5-2025-08-07
\item no\_none\_of\_the\_above: google/gemini-2.5-flash
\item no\_vague\_terms: google/gemini-2.5-flash
\item ordered\_options: google/gemini-2.5-pro
\item plausible\_distractors: openai/gpt-5-2025-08-07
\item problem\_in\_stem: google/gemini-2.5-flash-lite
\item single\_best\_answer': google/gemini-2.5-flash
\end{itemize}

\subsection{Annotation Details} \label{appendix:annotation}

Our human annotation protocol draws on qualitative coding in \textsc{hci} \cite{bingham2023data}: three authors label the three different metrics (one author per metric), then a second author rates 50 random items to compute agreement.
The annotators are graduate students in computer science with native fluency in English, all with previous experience in evaluation research and \mcq{} annotations.
Contamination yields 84\% agreement, shortcuts yields 84\% agreement, and the lowest agreement on any writing flaw is 85\%.
The annotation protocols for contamination, shortcuts, and writing flaws are in \cref{subsection:annotations}; the rule definitions shown for writing flaws are in Table~\ref{table:writing_flaw_rules}.
This annotation was deemed except by our institution's Internal Review Board (\textsc{irb}).

\subsection{Full Writing Flaw Results} \label{appendix:writing_flaw_results}

Due to space constraints, we report writing flaw scores aggregated over all flaws or with a standard cutoff of violating two or more rules, but we now analyze each individual rule.
In Figure~\ref{fig:writing_flaw_all_score}, we show the score of each writing flaw across all datasets, which provides interesting, dataset-specific quirks; AQuA and SuperGPQA fail to sort their choices, HellaSwag is the main culprit of using vague terms, and TruthfulQA is the only benchmark with pervasive convergence cues. 
Figure~\ref{fig:writing_flaw_all_prevalence} extends Figure~\ref{fig:writing_flaws} to~sort the prevalence of the 19 writing flaws across student-based and non-student-based benchmarks, which could inform education or \nlp{} researchers as to which issues are important to immediately fix.

\subsection{Full Search Engine Results} \label{appendix:search_engine_results}

Table~\ref{table:contamination_all} reports our contamination results across all judge and search engine combinations. Along with the GPT-5, Gemini-2.5 Pro, and Claude-4.5 Sonnet judges, we also assess: 1) an Oracle judge, which makes a perfect/oracle decision if any search results appear; and 2) a Simple judge, which classifies the \mcq{} as flawed if any search results appear, and not flawed if there are no search results.
Ignoring the Oracle classifier, Google is consistently the strongest search engine.
The simple classifier slightly surpasses GPT-5 at using Google's search results, but GPT-5 can also generate an explanation that synthesizes all of the web pages; we personally found this useful for debugging and expect other researchers to feel similarly, so we use GPT-5 for contamination detection in \model{}.

\subsection{Results with Item Response Theory} \label{appendix:irt_results}

Our analysis in \cref{subsection:impact} reports the average accuracy of 10 \mm{}s to study the impact of flaws in \mm{} evaluation---a proxy for the ``difficulty'' of an \mcq{}.
However, this measure could become more informative after considering model abilities.
To illustrate, an \mcq{} answered just by GPT-5 Nano and an \mcq{} answered just by GPT-5 would have the same accuracy, but the latter is more difficult, as we know GPT-5 tends to have higher accuracy on average \cite{sung2024your}.
Item Response Theory \cite[\textsc{irt}]{lord1964effect} is a tool from educational testing that controls for this; it estimates the difficulty (how hard the \mcq{} is) and discriminability (how well the \mcq{} discerns model skills) by learning the abilities of the models run on the \mcqa{} benchmarks.

While \textsc{irt} is a standard metric to report when validating education interventions \cite{schmucker2025impact}, and with recent growing interest in \nlp{} \cite{hofmann2025fluid}, we felt average accuracy would be more familiar and easier to interpret for an \nlp{} audience.
To reap the benefits of \textsc{irt}, we replicate the analysis in \cref{subsection:impact} but with average difficulty (Table~\ref{table:difficulty_flaw}) and discriminability (Table~\ref{table:disc_flaw}) as the metrics, which do not alter our claims; contaminated items have lower difficulty and discriminability, while items with shortcuts and writing errors have higher difficulty and discriminability.

\subsection{Contamination Detection in Pre-training} \label{appendix:compare_contamination}

Prior work offers many ways to predict whether test items exist in \mm{} training sets~\cite{fu-etal-2025-data}.
Some methods query black-box \mm{}s \cite{sainz-etal-2023-nlp}, but we desire a model-agnostic technique, isolating \mcq{} contamination irrespective of model behavior.
Other methods search large corpora, but these either rely on exact string matching \cite{xu-etal-2025-infini}---missing perturbed \mcq{}s---or index entire corpora \cite{elazar2024whats}---which is resource-intensive.
Instead, we use the Internet as a proxy for training data, assuming if an \mcq{} exists online, it likely exists in at least one \mm{}'s training data \cite{balloccu-etal-2024-leak}.
This is model-agnostic, cheap, and uses relevance ranking for near matches.

While search engine \textsc{api}s do not require intense resources and can surface semantic matches (\cref{subsection:contamination}), there are cheaper alternatives like Infini-Gram \cite{liu2024infini}: an $n$-gram (trillion) language model with publicly-available indexes over common pre-training corpora.
To see whether Infini-Gram can more efficiently replace our judges, we run the tool indexed over varied corpora on our contamination validation set---predicting the item as contaminated when the question stem exists exactly at least once.
Overall, accuracy and Cohen's $\kappa$ are significantly lower (Table~\ref{table:contamination_pretraining}), demonstrating that our method is a stronger way to detect whether \mcq{}s exist online.

\subsection{Shortcut Detection Across Models} \label{appendix:shortcut_ablation}

Our shortcut detection strategy uses majority vote of three strong \mm{}s that answer \mcq{}s without the question (\cref{subsection:shortcuts}), so we now test differences~when we use one \mm{}.
In Figure~\ref{fig:appendix:shortcut_scores} and Table~\ref{table:accuracy_by_shortcut}, we replicate Figure~\ref{fig:dataset_scores} and Table~\ref{table:accuracy_flaw}, evaluating the prevalence of shortcuts and their impact on \mm{} evaluation.
Trends are consistent with majority vote---relative shortcut prevalence is typically preserved across datasets and items with shortcuts have weakly lower accuracy---but there is model-specific noise, motivating the benefits of an ensembling approach.

In Table~\ref{table:choices_only}, we also study \mm{} accuracy changes when evaluating on \mcq{}s with and without choices-only success.
We see items where models succeed with choices-only have much higher accuracy---reproducing \citet{gupta2024improving}---which confirms that accounting for \textit{how} \mm{}s achieve choices-only success has substantially different implications on evaluation.
We also note that our choices-only accuracy aligns closely with \citet{balepur2025test}.

\subsection{\model{} Cost Analysis} \label{appendix:cost}

We report the input tokens consumed and output tokens produced in our \model{} runs across datasets in Table~\ref{appendix:table:cost}.
For each run, we also show the estimated cost per item using the pricing for Gemini-2.5 Flash and Pro.\footnote{https://ai.google.dev/gemini-api/docs/pricing}
While Gemini Pro is a slightly more reliable model (Table~\ref{table:combined_shortcuts_writing}), Gemini-2.5 Flash has similarly Cohen's $\kappa$ and significantly cheaper, suggesting the latter is a strong choices for researchers with limited computation budgets.

\subsection{Open-Weight Configurations} \label{appendix:open_weight}

To improve open-weight \mm{}s for writing flaw detection (\cref{subsection:recommendation}), we run experiments on ensembling, confidence, and writing flaw failure taxonomies.

\paragraph{Ensembling.} To test benefits of ensembling, we test all combinations of three open-weight judges; Command-R, Command-R+, and Qwen3-8B reach the highest Cohen's $\kappa$ of 0.391.
However, this is not much higher than Command-R alone ($\kappa$ of 0.376), so ensembling is likely not worth this extra computation for writing flaw detection.

\paragraph{Confidence.} To study the benefits of confidence calibration, we analyze the 1--10 confidence scores produced by LLM judges in Prompt~\ref{prompt:writing_flaw}.
In Table~\ref{appendix:table:confidence}, we show Cohen's $\kappa$ and the proportion of examples predicted (in parentheses) beyond different confidence thresholds.
Overall, some open-weight models are well-calibrated, with Command R reaching Cohen’s Kappa as high as 0.67 when giving a confidence score of 10.

\paragraph{Failure Taxonomy.} The above analyses and~results in Table~\ref{table:combined_shortcuts_writing} support that Cohere Command-R is a strong open-weight model for writing error detection, so we provide a breakdown of its Cohen's $\kappa$ across the 19 flaw types in Table~\ref{appendix:table:flaw_breakdown} to see where the model fails and succeeds.
Command-R excels in simpler judgments like detecting certain options and option order, but disagrees with experts more on complex, subjective criteria like extraneousness, repetition, and convergence.

\paragraph{Recommendation.} Cohere Command-R has the highest agreement with humans in Table~\ref{appendix:table:flaw_breakdown}, is part of the strongest judge ensemble, and appears well-calibrated, so we recommend that researchers with GPUs only use Command-R.
If some API credits are available, we recommend researchers to employ stronger closed-source \mm{}s when Command-R predicts confidence lower than 9, and for writing error types the model tends to struggle with, like convergent clues, extraneous info, and repetition.

\subsection{Prompts} \label{appendix:prompts}

This section outlines our prompts.
The prompt we use for running \mm{}s on \mcqa{} (Prompt~\ref{prompt:mcqa}). is taken directly from Inspect \cite{inspect_ai_framework}, except we do not ask the model's answer to be preceded by the dollar sign character (\$); preliminary analysis found many \mm{}s struggled with this, lowering scores more than normal.
Prompts~\ref{prompt:contamination}, \ref{prompt:choices_only}, \ref{prompt:question_similarity}, and \ref{prompt:writing_flaw} are the \mm{} instructions for detecting contaminated \mcq{}s on web pages, using \mm{}s to answer \mcq{}s with just the choices, detecting whether the inferred question matches the original one, and following the 19-rule education rubric for writing flaws, respectively.
For contamination, we consider the item to be flawed if the \mm{} judge outputs ``partial\_match'' or ``no\_match'', and for shortcuts, we consider the item flawed if the \mm{} judge outputs ``no\_match''.

\clearpage

\input{appendix/writing_flaw_rules}

\input{appendix/inspect}
\input{appendix/diff_discrim}

\input{appendix/all_writing_flaws}
\input{appendix/all_contamination}
\input{appendix/pretraining_contamination}
\input{appendix/shortcut_ablation}
\input{appendix/cost}
\input{appendix/open_weight}
\clearpage
\input{appendix/prompts}

%% file: appendix/writing_flaw_rules.tex
\newcommand{\redhl}[1]{\colorbox{red!25}{#1}}

\begin{table*}[]
\small
\centering
\renewcommand{\arraystretch}{0.9}
\setlength{\tabcolsep}{2pt}
\begin{tabular}{@{}lll@{}}
\toprule
\textbf{Name} & \textbf{Rule} & \textbf{Example Violation} \\
\midrule

Grammatical Consistency &
  \specialcellleft{All options must use parallel\\grammatical structure and fit with the stem.} &
  \begin{tabular}[c]{@{}l@{}}The moon orbits an object that orbits \redhl{the}\\ (D) \redhl{mars}\end{tabular} \\
\midrule

Focused Stem &
  \specialcellleft{The stem must present one clear,\\focused question or problem.} &
  \redhl{What are social?} \\
\midrule

Problem in Stem &
  \specialcellleft{The stem must fully state the problem\\ rather than relying on the options to introduce it.} &
  \specialcellleft{\redhl{Winter in the Northern Hemisphere means}} \\
\midrule

Single Best Answer &
  \specialcellleft{The question must have exactly one best\\ answer with no equally correct alternatives.} &
  \begin{tabular}[c]{@{}l@{}}Where could you get something that is \\made out of wool but cannot be worn?\\(C) \redhl{fabric store}\\ (E) \redhl{clothing factory}\end{tabular} \\
\midrule

No Extraneous Information &
  \specialcellleft{The stem and choices must exclude text\\not needed to answer the question.} &
  \begin{tabular}[c]{@{}l@{}}After school, Alex took Sasha's daughter\\to the playground to play. \redhl{The two were}\\ \redhl{good friends in the same class}.\\ What will Sasha want to do next?\end{tabular} \\
\midrule

Avoid K-Type Options &
  \specialcellleft{Choices must not\\combine items (e.g., “A and B only”)}. &
  \begin{tabular}[c]{@{}l@{}}(A) \redhl{Wrong, Wrong}\\ (B) \redhl{Wrong, Not wrong}\\ (C) \redhl{Not wrong, Wrong}\\ (D) \redhl{Not wrong, Not wrong}\end{tabular} \\
\midrule

Clear Language &
  All wording must be clear and unambiguous. &
  \specialcellleft{What lasts only as long as\\ \redhl{the antibodies survive in body fluids}?} \\
\midrule

Plausible Distractors &
  \specialcellleft{All distractors must be plausible\\and relevant to the topic in the question stem.} &
  \begin{tabular}[c]{@{}l@{}}[math question asking for \redhl{probability}]\\ (E) \redhl{1.5}\end{tabular} \\
\midrule

Avoid Repetition &
  \specialcellleft{The correct answer must not repeat words\\ or phrases from the question stem.} &
  \begin{tabular}[c]{@{}l@{}}In the morning you return to \redhl{work}, in the evening you?\\ Answer: leave \redhl{work}\end{tabular} \\
\midrule

No Logical Cues &
  \specialcellleft{Choices must not provide logical cues\\that reveal the correct answer.} &
  \begin{tabular}[c]{@{}l@{}}Who has blood and \redhl{parents}?\\ (A) \redhl{person}\\ (B) bloodbank\\ (C) vein\\ (D) capillaries\\ (E) hospital\end{tabular} \\
\midrule

No Convergence Cues &
  \specialcellleft{The correct answer must not combine elements\\repeatedly appearing across distractors.} &
  \begin{tabular}[c]{@{}l@{}}(A) an \redhl{abundance} of fire\\ (B) absolutely zero \redhl{snow} outside\\ (C) a \redhl{plethora} of \redhl{snow} (answer)\\ (D) frogs falling from sky\end{tabular} \\
\midrule

Equal-Length Options &
  All choices must have similar length and detail. &
  \begin{tabular}[c]{@{}l@{}}(A) \redhl{a violent disturbance of the atmosphere}\\\redhl{with high winds}\\ (B) ideas in print media\\ (C) a large jet engine\\ (D) a gas guzzling automobile\end{tabular} \\
\midrule

Ordered Options &
  Numerical choices must be in ascending order. &
  \begin{tabular}[c]{@{}l@{}}(A) 18cm\\ (B) \redhl{22cm}\\ (C) \redhl{20cm}\\ (D) 30cm\\ (E) 28cm\end{tabular} \\
\midrule

No Absolute Terms &
  \specialcellleft{Choices must not use absolute terms\\unless the statement is truly absolute.} &
  \begin{tabular}[c]{@{}l@{}}Will happen to the number of islands if\\the planet's temperature rises?\\ (A) they will increase\\ (B) \redhl{nothing will happen}\\ (C) they will shrink\\ (D) they will double\end{tabular} \\
\midrule

No Vague Terms &
  \specialcellleft{Choices must not use vague,\\unquantified terms like “often” or “usually.”} &
  \begin{tabular}[c]{@{}l@{}}When a water balloon is frozen, it will contain\\ (A) \redhl{a much less amount} of water\\
(B) \redhl{a whole bunch} of frozen ice-cream\\ \end{tabular} \\
\midrule

Avoid Negative Stems &
  \specialcellleft{The stem must not be framed negatively\\ using terms like “NOT” or “EXCEPT.”} &
  \specialcellleft{All of the following are ways in which\\lobbyists attempt to persuade legislators \redhl{EXCEPT}...} \\
\midrule

No “All of the Above” &
  Choices should not include “All of the above.” &
  \begin{tabular}[c]{@{}l@{}}(D) \redhl{all of these}\end{tabular} \\
\midrule

No “None of the Above” &
  Choices should not include “None of the above” &
  \begin{tabular}[c]{@{}l@{}}(E) \redhl{None of these}\end{tabular} \\
\midrule

No Fill-in-the-Blank &
  \specialcellleft{The stem must not use blanks or require\\choices to complete an incomplete sentence.} &
  Experiments are performed in the \redhl{\_\_\_}. \\
\bottomrule
\end{tabular}
\caption{\label{table:writing_flaw_rules} The 19 writing flaw rubric from \cite{tarrant2006frequency} and example violations from \mcqa{} benchmarks.}
\end{table*}

%% file: appendix/inspect.tex
\begin{figure*}
    \centering
    \fbox{\includegraphics[width=\linewidth]{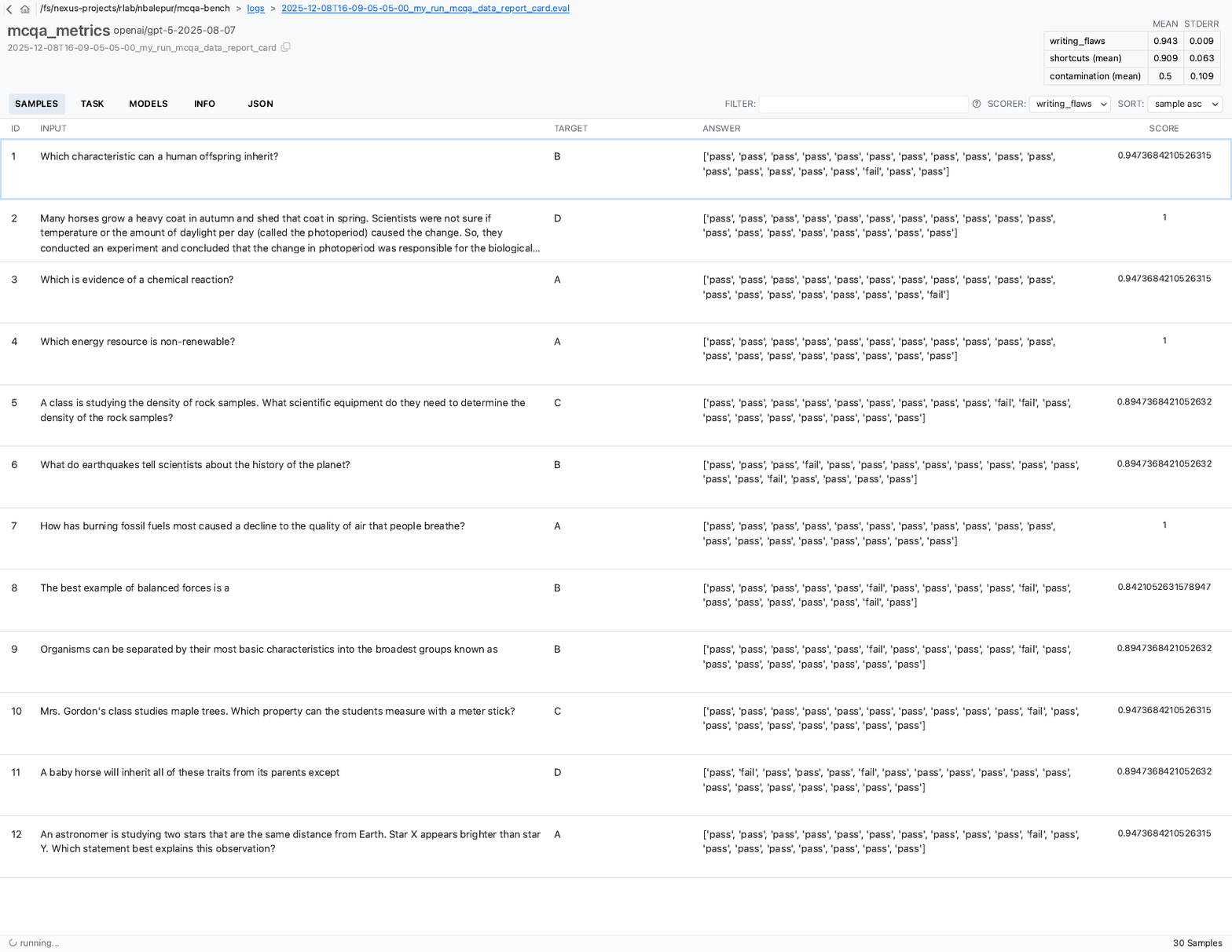}}
    \caption{ Interface from InspectAI for viewing \model{} runs. The overall scores are in the top right, and researchers can click on specific \mcq{}s to view \mm{} judge calls and feedback, supporting debugging and analysis.}
    \label{fig:appendix:inspect}
\end{figure*}

%% file: appendix/diff_discrim.tex
\begin{table*}
\centering
\scriptsize
\setlength{\tabcolsep}{2.5pt}

\begin{tabular}{l|cccc|cccc|cccc|cccc}
\multicolumn{1}{c}{} & \multicolumn{4}{c}{\textit{Contamination}} & \multicolumn{4}{c}{\textit{Shortcuts}} & \multicolumn{4}{c}{\textit{2+ Writing Flaws}} & \multicolumn{4}{c}{\textit{Any Flaw}} \\
\toprule
Dataset & Flaw & No Flaw & $\Delta$Acc & $\probP$(Flaw) & Flaw & No Flaw & $\Delta$Acc & $\probP$(Flaw) & Flaw & No Flaw & $\Delta$Acc & $\probP$(Flaw) & Flaw & No Flaw & $\Delta$Acc & $\probP$(Flaw) \\
\midrule
AQUA & 0.007 & -0.046 & \negative{-742.1} & 17\% & -0.032 & -0.038 & \positive{+19.3} & 20\% & 0.236 & -0.252 & \negative{-206.8} & 44\% & 0.069 & -0.218 & \negative{-414.0} & 63\% \\
ARC & -0.616 & -0.493 & \negative{-19.9} & 38\% & -0.410 & -0.548 & \positive{+33.8} & 6\% & -0.468 & -0.595 & \positive{+27.1} & 43\% & -0.541 & -0.539 & \negative{-0.4} & 70\% \\
CQA & -0.096 & -0.217 & \positive{+126.0} & 6\% & 0.092 & -0.221 & \negative{-341.8} & 4\% & -0.193 & -0.638 & \positive{+230.2} & 96\% & -0.196 & -0.624 & \positive{+217.9} & 97\% \\
HS & -0.501 & -0.181 & \negative{-63.8} & 0\% & -0.008 & -0.212 & \positive{+2666.1} & 14\% & -0.182 & --- & --- & 100\% & -0.182 & --- & --- & 100\% \\
MMLU & -0.420 & -0.055 & \negative{-86.9} & 45\% & -0.102 & -0.232 & \positive{+127.7} & 9\% & -0.135 & -0.372 & \positive{+175.1} & 64\% & -0.223 & -0.204 & \negative{-8.7} & 84\% \\
OBQA & -0.501 & -0.553 & \positive{+10.3} & 4\% & -0.513 & -0.561 & \positive{+9.3} & 21\% & -0.530 & -0.734 & \positive{+38.3} & 90\% & -0.534 & -0.749 & \positive{+40.4} & 92\% \\
PIQA & -0.690 & -0.725 & \positive{+5.0} & 5\% & -0.543 & -0.770 & \positive{+41.9} & 20\% & -0.714 & -0.834 & \positive{+16.8} & 92\% & -0.715 & -0.848 & \positive{+18.6} & 94\% \\
QASC & 0.491 & 0.472 & \negative{-3.9} & 6\% & 0.639 & 0.466 & \negative{-27.1} & 4\% & 0.485 & 0.082 & \negative{-83.2} & 97\% & 0.479 & 0.257 & \negative{-46.2} & 98\% \\
SAT & 0.073 & -0.120 & \negative{-264.9} & 24\% & -0.072 & -0.075 & \positive{+4.6} & 21\% & 0.033 & -0.111 & \negative{-440.3} & 25\% & -0.040 & -0.115 & \positive{+188.1} & 54\% \\
SIQA & -1.096 & -0.331 & \negative{-69.8} & 0\% & -0.326 & -0.333 & \positive{+2.2} & 23\% & -0.298 & -0.993 & \positive{+233.0} & 95\% & -0.304 & -1.004 & \positive{+230.7} & 96\% \\
SGPQA & 1.006 & 0.992 & \negative{-1.4} & 6\% & 0.831 & 1.011 & \positive{+21.6} & 10\% & 1.082 & 0.687 & \negative{-36.5} & 77\% & 1.064 & 0.691 & \negative{-35.1} & 81\% \\
TQA & -0.123 & -0.181 & \positive{+47.5} & 47\% & 0.543 & -0.195 & \negative{-135.9} & 6\% & -0.133 & -0.823 & \positive{+516.8} & 97\% & -0.145 & -0.863 & \positive{+495.9} & 99\% \\
\midrule
$\text{Micro } \mu$ & -0.312 & -0.114 & \negative{-63.4} & 14\% & -0.172 & -0.138 & \negative{-20.0} & 12\% & -0.101 & -0.348 & \positive{+245.6} & 83\% & -0.129 & -0.250 & \positive{+93.1} & 90\% \\
$\text{Macro } \mu$ & -0.205 & -0.120 & \negative{-41.7} & 17\% & 0.008 & -0.142 & \negative{-1816.3} & 13\% & -0.068 & -0.417 & \positive{+510.6} & 77\% & -0.106 & -0.383 & \positive{+262.9} & 85\% \\
\bottomrule
\end{tabular}
\caption{Impact of \mcqa{} benchmark flaws on \mm{} difficulty, computed via Item Response Theory. The trend is consistent with Table~\ref{table:accuracy_flaw}; contaminated items have lower difficulty, while items with shortcuts and writing flaws have higher difficulty.  \label{table:difficulty_flaw}}
\end{table*}

\begin{table*}
\centering
\scriptsize
\setlength{\tabcolsep}{2.5pt}
\begin{tabular}{l|cccc|cccc|cccc|cccc}
\multicolumn{1}{c}{} & \multicolumn{4}{c}{\textit{Contamination}} & \multicolumn{4}{c}{\textit{Shortcuts}} & \multicolumn{4}{c}{\textit{2+ Writing Flaws}} & \multicolumn{4}{c}{\textit{Any Flaw}} \\
\toprule
Dataset & Flaw & No Flaw & $\Delta$Acc & $\probP$(Flaw) & Flaw & No Flaw & $\Delta$Acc & $\probP$(Flaw) & Flaw & No Flaw & $\Delta$Acc & $\probP$(Flaw) & Flaw & No Flaw & $\Delta$Acc & $\probP$(Flaw) \\
\midrule
AQUA & 1.282 & 1.249 & \negative{-2.5} & 17\% & 1.236 & 1.259 & \positive{+1.9} & 20\% & 1.217 & 1.285 & \positive{+5.5} & 44\% & 1.232 & 1.294 & \positive{+5.1} & 63\% \\
ARC & 1.384 & 1.361 & \negative{-1.6} & 38\% & 1.295 & 1.374 & \positive{+6.1} & 6\% & 1.376 & 1.366 & \negative{-0.7} & 43\% & 1.377 & 1.352 & \negative{-1.8} & 70\% \\
CQA & 1.295 & 1.254 & \negative{-3.1} & 6\% & 1.169 & 1.260 & \positive{+7.8} & 4\% & 1.255 & 1.316 & \positive{+4.9} & 96\% & 1.255 & 1.315 & \positive{+4.7} & 97\% \\
HS & 1.075 & 1.235 & \positive{+14.9} & 0\% & 1.167 & 1.246 & \positive{+6.8} & 14\% & 1.235 & --- & --- & 100\% & 1.235 & --- & --- & 100\% \\
MMLU & 1.343 & 1.252 & \negative{-6.7} & 45\% & 1.256 & 1.297 & \positive{+3.3} & 9\% & 1.282 & 1.313 & \positive{+2.4} & 64\% & 1.300 & 1.258 & \negative{-3.2} & 84\% \\
OBQA & 1.297 & 1.325 & \positive{+2.2} & 4\% & 1.325 & 1.324 & \negative{-0.1} & 21\% & 1.317 & 1.386 & \positive{+5.3} & 90\% & 1.317 & 1.399 & \positive{+6.2} & 92\% \\
PIQA & 1.357 & 1.321 & \negative{-2.6} & 5\% & 1.284 & 1.333 & \positive{+3.8} & 20\% & 1.321 & 1.353 & \positive{+2.4} & 92\% & 1.321 & 1.353 & \positive{+2.4} & 94\% \\
QASC & 1.208 & 1.256 & \positive{+4.0} & 6\% & 1.174 & 1.257 & \positive{+7.0} & 4\% & 1.252 & 1.296 & \positive{+3.5} & 97\% & 1.253 & 1.264 & \positive{+0.9} & 98\% \\
SAT & 1.374 & 1.352 & \negative{-1.5} & 24\% & 1.357 & 1.358 & \positive{+0.1} & 21\% & 1.330 & 1.367 & \positive{+2.8} & 25\% & 1.347 & 1.369 & \positive{+1.7} & 54\% \\
SIQA & 1.485 & 1.252 & \negative{-15.7} & 0\% & 1.241 & 1.256 & \positive{+1.2} & 23\% & 1.244 & 1.406 & \positive{+13.0} & 95\% & 1.246 & 1.411 & \positive{+13.3} & 96\% \\
SGPQA & 1.160 & 1.221 & \positive{+5.2} & 6\% & 1.217 & 1.217 & \positive{+0.0} & 10\% & 1.214 & 1.226 & \positive{+0.9} & 77\% & 1.216 & 1.220 & \positive{+0.3} & 81\% \\
TQA & 1.256 & 1.280 & \positive{+1.9} & 47\% & 1.211 & 1.272 & \positive{+5.0} & 6\% & 1.268 & 1.302 & \positive{+2.7} & 97\% & 1.269 & 1.267 & \negative{-0.2} & 99\% \\
\midrule
$\text{Micro } \mu$ & 1.324 & 1.273 & \negative{-3.8} & 14\% & 1.253 & 1.284 & \positive{+2.5} & 12\% & 1.271 & 1.330 & \positive{+4.6} & 83\% & 1.277 & 1.312 & \positive{+2.7} & 90\% \\
$\text{Macro } \mu$ & 1.293 & 1.280 & \negative{-1.0} & 17\% & 1.244 & 1.288 & \positive{+3.5} & 13\% & 1.276 & 1.329 & \positive{+4.1} & 77\% & 1.281 & 1.318 & \positive{+2.9} & 85\% \\
\bottomrule
\end{tabular}
\caption{Impact of \mcqa{} benchmark flaws on \mm{} discriminability, computed via Item Response Theory. The trend is consistent with Table~\ref{table:accuracy_flaw}; contaminated items have lower discriminability, while items with shortcuts and writing flaws have higher discriminability.  \label{table:disc_flaw}}
\end{table*}

%% file: appendix/all_writing_flaws.tex
\begin{figure*}
    \centering
    \includegraphics[width=\linewidth]{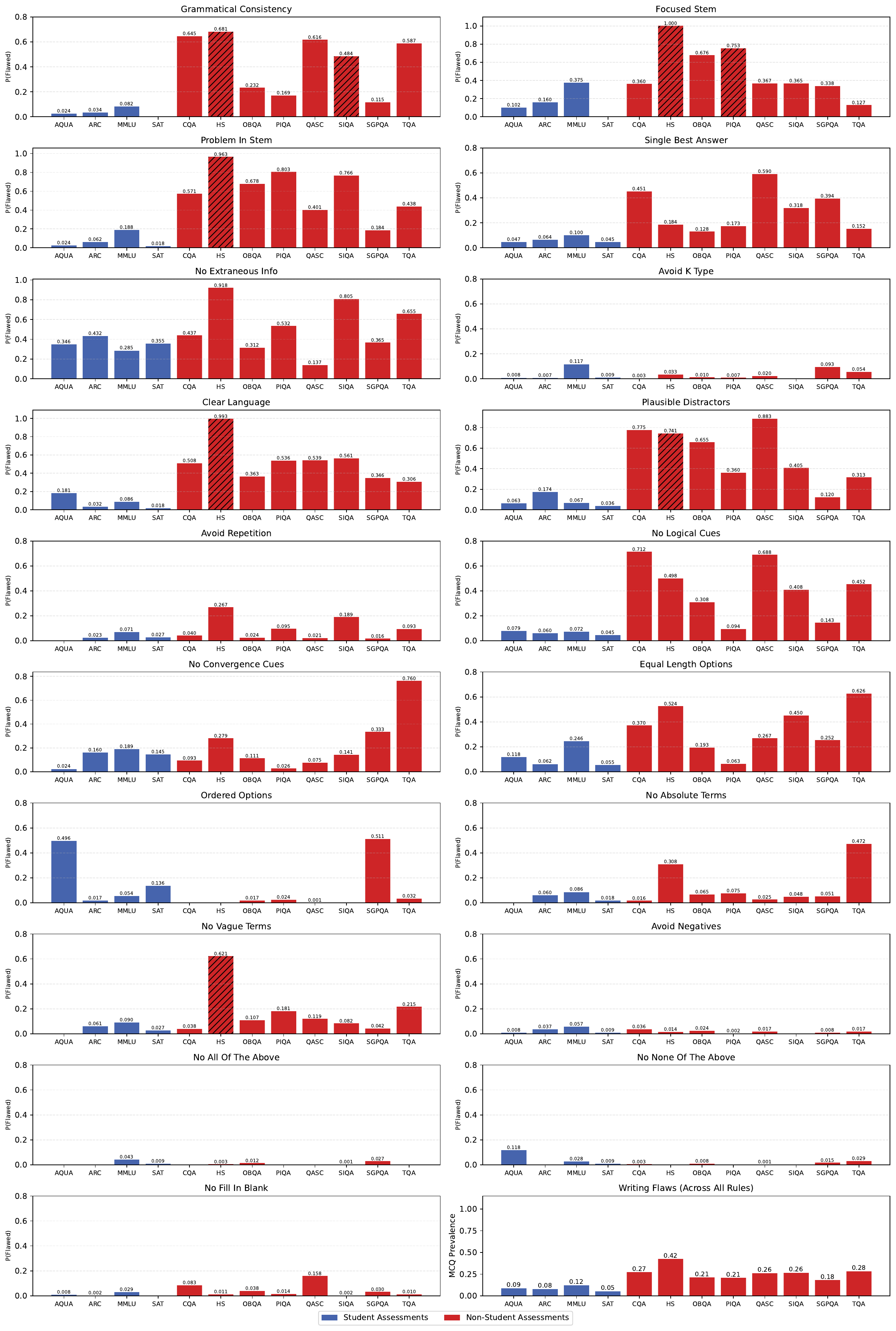}
    \caption{ Scores for each of the 19 writing flaws across each \mcqa{} benchmark.}
    \label{fig:writing_flaw_all_score}
\end{figure*}

\begin{figure*}
    \centering
    \includegraphics[width=\linewidth]{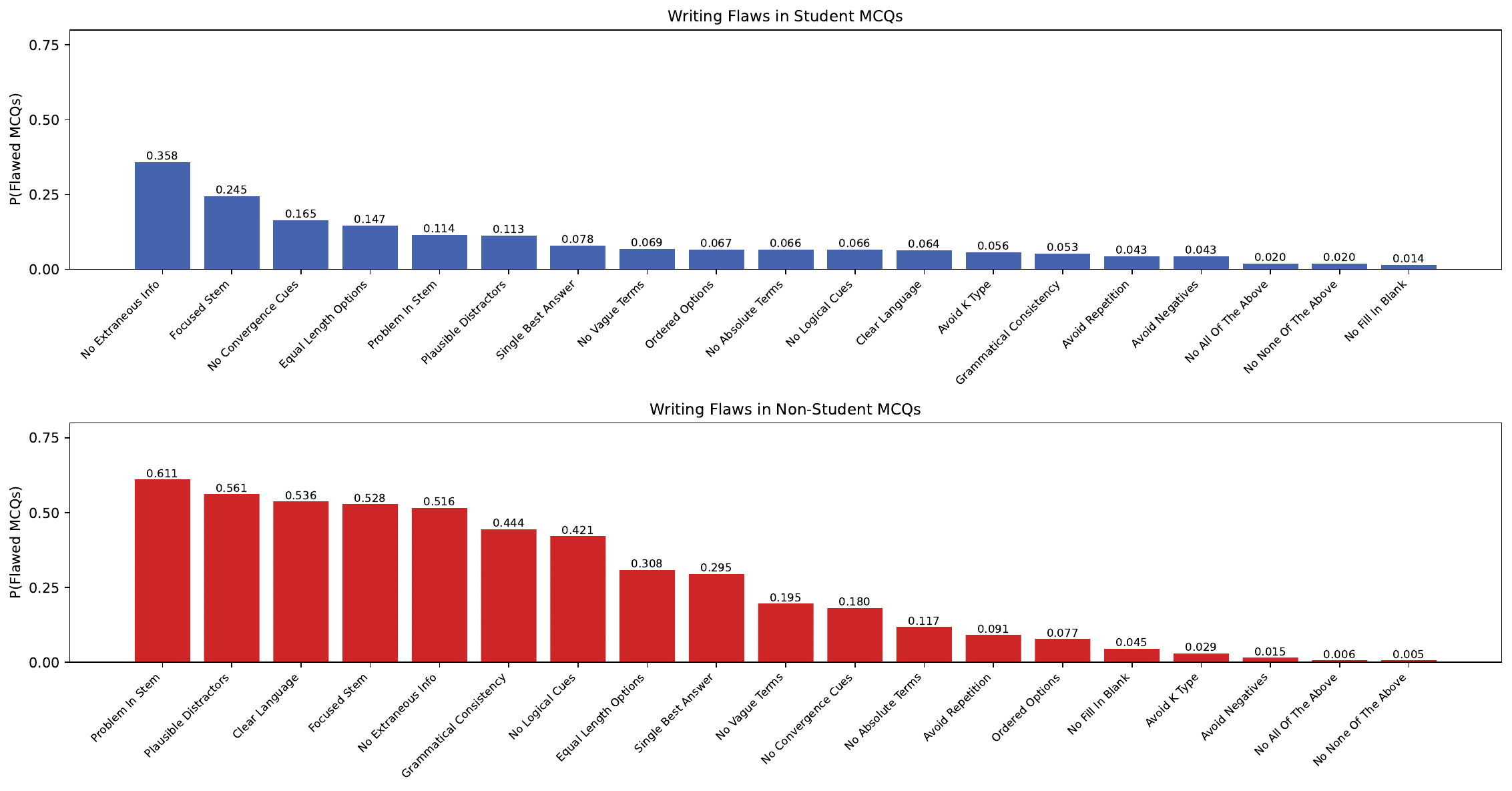}
    \caption{ Descending prevalence of all $19$ writing flaws across exam-based and non-exam-based \mcqa{} benchmarks.}
    \label{fig:writing_flaw_all_prevalence}
\end{figure*}

%% file: appendix/all_contamination.tex
\begin{table*}[]
\centering
\begin{tabular}{@{}llcccc@{}}
\toprule
Judge Classification Type & Search Engine & Accuracy & F1 Score & Cohen's $\kappa$ \\ \midrule
Oracle & Google & 0.8035 & 0.7761 & 0.6161 \\
Oracle & Brave & 0.6026 & 0.4129 & 0.2456 \\
Oracle & Perplexity & 1.0000 & 1.0000 & 1.0000 \\
Oracle & Exa & 0.9825 & 0.9835 & 0.9650 \\
Oracle & Tavily & 0.9694 & 0.9707 & 0.9388 \\
Oracle & Serper & 0.7598 & 0.7120 & 0.5337 \\
\midrule
Simple & Google & 0.7293 & 0.7156 & 0.4653 \\
Simple & Brave & 0.5852 & 0.4025 & 0.2105 \\
Simple & Perplexity & 0.5371 & 0.6989 & 0.0000 \\
Simple & Exa & 0.5546 & 0.7000 & 0.0458 \\
Simple & Tavily & 0.5197 & 0.6784 & -0.0305 \\
Simple & Serper & 0.6463 & 0.6267 & 0.3019 \\
\midrule
GPT-5 & Google & 0.7118 & 0.6765 & 0.4358 \\
GPT-5 & Brave & 0.5415 & 0.2759 & 0.1349 \\
GPT-5 & Perplexity & 0.6419 & 0.5287 & 0.3122 \\
GPT-5 & Exa & 0.5895 & 0.4268 & 0.2164 \\
GPT-5 & Tavily & 0.5808 & 0.4074 & 0.2007 \\
GPT-5 & Serper & 0.6725 & 0.6445 & 0.3560 \\
\midrule
Claude Sonnet & Google & 0.6900 & 0.6359 & 0.3965 \\
Claude Sonnet & Brave & 0.5415 & 0.2657 & 0.1359 \\
Claude Sonnet & Perplexity & 0.6332 & 0.5385 & 0.2919 \\
Claude Sonnet & Exa & 0.5764 & 0.4049 & 0.1919 \\
Claude Sonnet & Tavily & 0.5808 & 0.4000 & 0.2017 \\
Claude Sonnet & Serper & 0.6507 & 0.6226 & 0.3127 \\
\midrule
Gemini Pro & Google & 0.6987 & 0.6497 & 0.4128 \\
Gemini Pro & Brave & 0.5415 & 0.2657 & 0.1359 \\
Gemini Pro & Perplexity & 0.6288 & 0.5304 & 0.2839 \\
Gemini Pro & Exa & 0.5764 & 0.4049 & 0.1919 \\
Gemini Pro & Tavily & 0.5983 & 0.4321 & 0.2340 \\
Gemini Pro & Serper & 0.6638 & 0.6351 & 0.3389 \\
\bottomrule
\end{tabular}
\caption{Contamination detection results across all judge and search engine combinations.}
\label{table:contamination_all}
\end{table*}

%% file: appendix/pretraining_contamination.tex
\begin{table*}[]
\centering
\begin{tabular}{@{}lccc@{}}
\toprule
Pretraining Corpus & Accuracy & F1 Score & Cohen's $\kappa$ \\ \midrule
\texttt{v4\_olmo-2-0325-32b-instruct\_llama} & 0.5590 & 0.6576 & 0.0837 \\
\texttt{v4\_dclm-baseline\_llama} & 0.5153 & 0.6626 & -0.0306 \\
\texttt{v4\_dolma-v1\_7\_llama} & 0.5284 & 0.5814 & 0.0441 \\
\texttt{v4\_rpj\_llama\_s4} & 0.5415 & 0.6828 & 0.0238 \\
\texttt{v4\_piletrain\_llama} & 0.5459 & 0.6959 & 0.0257 \\
\texttt{v4\_c4train\_llama} & 0.5633 & 0.7059 & 0.0645 \\
\bottomrule
\end{tabular}
\caption{\small Contamination detection agreement with human judgments when using exact question matches in pretraining corpora via Infini-Gram \cite{liu2024infini}. All methods have much lower accuracy and Cohen's $\kappa$ than \mm{} judges with search \textsc{api}s.}
\label{table:contamination_pretraining}
\end{table*}

%% file: appendix/shortcut_ablation.tex
\begin{figure*}
    \centering
    \includegraphics[width=\linewidth]{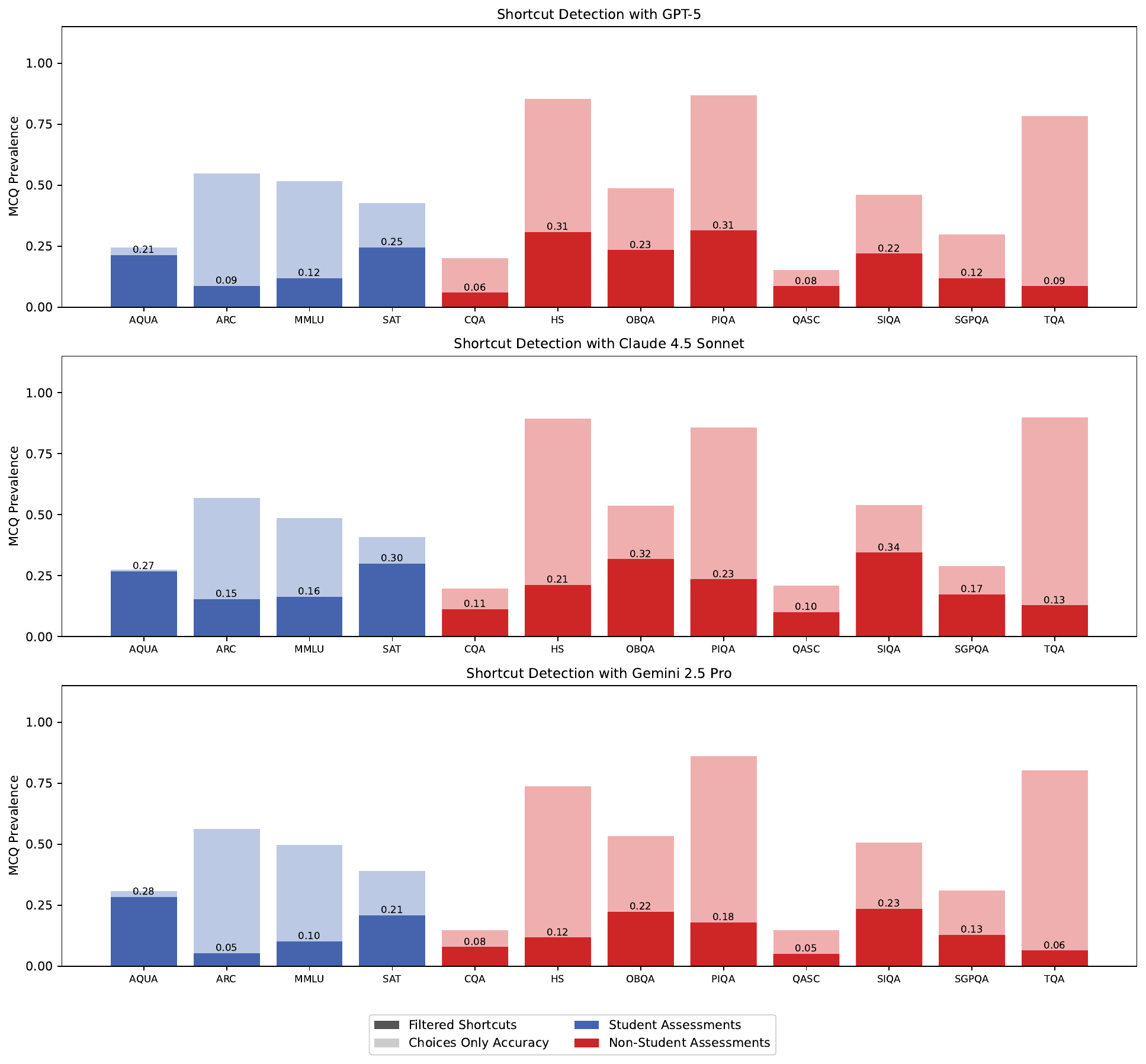}
    \caption{\small Shortcut prevalence on \mcq{} benchmarks with different models, depending on the model used to answer the \mcq{} with just the choices. Overall trends are relatively consistent, but there are model-specific differences, motivating our design choices of taking majority vote over these three \mm{}s.}
    \label{fig:appendix:shortcut_scores}
\end{figure*}

\begin{table*}
\centering
\small
\setlength{\tabcolsep}{3.25pt}
\begin{tabular}{l|cccc|cccc|cccc}
\multicolumn{1}{c}{} & \multicolumn{4}{c}{\textit{GPT-5}} & \multicolumn{4}{c}{\textit{Claude 4.5 Sonnet}} & \multicolumn{4}{c}{\textit{Gemini 2.5 Pro}} \\
\toprule
Dataset & Flaw & No Flaw & $\Delta$Acc & $\probP$(Flaw) & Flaw & No Flaw & $\Delta$Acc & $\probP$(Flaw) & Flaw & No Flaw & $\Delta$Acc & $\probP$(Flaw) \\
\midrule
AQUA & 0.737 & 0.761 & \positive{+3.3} & 21\% & 0.771 & 0.751 & \negative{-2.6} & 27\% & 0.789 & 0.743 & \negative{-5.8} & 28\% \\
ARC & 0.835 & 0.883 & \positive{+5.7} & 9\% & 0.862 & 0.882 & \positive{+2.3} & 15\% & 0.793 & 0.884 & \positive{+11.5} & 5\% \\
CQA & 0.659 & 0.785 & \positive{+19.0} & 6\% & 0.771 & 0.778 & \positive{+1.0} & 11\% & 0.779 & 0.777 & \negative{-0.3} & 8\% \\
HS & 0.770 & 0.787 & \positive{+2.1} & 31\% & 0.739 & 0.793 & \positive{+7.3} & 21\% & 0.744 & 0.787 & \positive{+5.8} & 12\% \\
MMLU & 0.743 & 0.801 & \positive{+7.7} & 12\% & 0.773 & 0.798 & \positive{+3.2} & 16\% & 0.742 & 0.800 & \positive{+7.8} & 10\% \\
OBQA & 0.848 & 0.877 & \positive{+3.4} & 23\% & 0.879 & 0.866 & \negative{-1.5} & 32\% & 0.850 & 0.876 & \positive{+3.0} & 22\% \\
PIQA & 0.874 & 0.921 & \positive{+5.3} & 31\% & 0.873 & 0.916 & \positive{+5.0} & 23\% & 0.874 & 0.913 & \positive{+4.5} & 18\% \\
QASC & 0.547 & 0.608 & \positive{+11.2} & 8\% & 0.596 & 0.604 & \positive{+1.3} & 10\% & 0.504 & 0.608 & \positive{+20.7} & 5\% \\
SAT & 0.774 & 0.782 & \positive{+1.0} & 25\% & 0.800 & 0.771 & \negative{-3.6} & 30\% & 0.770 & 0.783 & \positive{+1.7} & 21\% \\
SIQA & 0.749 & 0.816 & \positive{+9.0} & 22\% & 0.833 & 0.785 & \negative{-5.7} & 34\% & 0.809 & 0.799 & \negative{-1.3} & 23\% \\
SGPQA & 0.524 & 0.473 & \negative{-9.6} & 12\% & 0.506 & 0.474 & \negative{-6.3} & 17\% & 0.507 & 0.475 & \negative{-6.3} & 13\% \\
TQA & 0.723 & 0.778 & \positive{+7.6} & 9\% & 0.669 & 0.788 & \positive{+17.7} & 13\% & 0.596 & 0.785 & \positive{+31.7} & 6\% \\
\midrule
\textbf{Micro-Average} & 0.765 & 0.766 & \positive{+0.2} & 17\% & 0.781 & 0.762 & \negative{-2.4} & 20\% & 0.763 & 0.767 & \positive{+0.5} & 13\% \\
\textbf{Macro-Average} & 0.732 & 0.773 & \positive{+5.5} & 17\% & 0.756 & 0.767 & \positive{+1.5} & 21\% & 0.730 & 0.769 & \positive{+5.4} & 14\% \\
\bottomrule
\end{tabular}
\caption{\small \label{table:accuracy_by_shortcut} Accuracy on flawed \mcq{}s with shortcuts and not flawed \mcq{}s without them, across varied choices-only models. The trend is relatively consistent per model: items with shortcuts tend to have lower accuracy, but the effect is relatively small.}
\end{table*}

\begin{table*}
\centering
\begin{tabular}{l|cccc}
\multicolumn{1}{c}{} & \multicolumn{4}{c}{\textit{Choices-Only Success}} \\
\toprule
Dataset & Flaw & No Flaw & $\Delta$Acc & $\probP$(Flaw) \\
\midrule
AQUA & 0.746 & 0.759 & \positive{+1.6} & 22\% \\
ARC & 0.899 & 0.853 & \negative{-5.1} & 55\% \\
CQA & 0.840 & 0.767 & \negative{-8.7} & 14\% \\
HS & 0.810 & 0.630 & \negative{-22.2} & 84\% \\
MMLU & 0.823 & 0.765 & \negative{-7.1} & 49\% \\
OBQA & 0.903 & 0.836 & \negative{-7.4} & 51\% \\
PIQA & 0.923 & 0.788 & \negative{-14.6} & 88\% \\
QASC & 0.758 & 0.580 & \negative{-23.5} & 13\% \\
SAT & 0.786 & 0.776 & \negative{-1.2} & 38\% \\
SIQA & 0.867 & 0.731 & \negative{-15.6} & 52\% \\
SGPQA & 0.558 & 0.450 & \negative{-19.3} & 27\% \\
TQA & 0.829 & 0.499 & \negative{-39.9} & 83\% \\
\midrule
\textbf{Micro-Average} & 0.845 & 0.690 & \negative{-18.3} & 49\% \\
\textbf{Macro-Average} & 0.812 & 0.703 & \negative{-13.4} & 48\% \\
\bottomrule
\end{tabular}
\caption{\label{table:choices_only} Accuracy when shortcut flaws are defined by choices only success, without analyzing inferred questions. We reproduce the results of \citet{gupta2024improving}: filtering \mcq{}s with choices-only success lowers benchmark scores.}
\end{table*}

%% file: appendix/cost.tex
\begin{table*}[h]
\small
\centering
\begin{tabular}{lcccc}
\toprule
Dataset & Tokens In / Item & Tokens Out / Item & Gemini 2.5 Flash Cost / Item (USD) & Gemini 2.5 Pro Cost / Item (\$) \\
\midrule
AQUA  & 18949 & 6623 & 0.02 & 0.12 \\
ARC   & 18836 & 5337 & 0.02 & 0.10 \\
CQA   & 18109 & 5675 & 0.02 & 0.10 \\
HS    & 20584 & 7340 & 0.02 & 0.13 \\
MMLU  & 19549 & 5848 & 0.02 & 0.11 \\
OBQA  & 18182 & 5515 & 0.02 & 0.10 \\
PIQA  & 18558 & 4916 & 0.02 & 0.10 \\
QASC  & 18333 & 7268 & 0.02 & 0.12 \\
SAT   & 19592 & 6251 & 0.02 & 0.11 \\
SIQA  & 18252 & 5326 & 0.02 & 0.10 \\
SGPQA & 22300 & 8951 & 0.03 & 0.15 \\
TQA   & 19403 & 6189 & 0.02 & 0.11 \\
\bottomrule
\end{tabular}
\caption{Token usage and estimated per-item cost across datasets for Gemini 2.5 Flash and Pro models.}
\label{appendix:table:cost}
\end{table*}

%% file: appendix/open_weight.tex
\begin{table*}[h]
\centering
\begin{tabular}{lcccc}
\toprule
Model & All & Confidence $\geq 8$ & Confidence $\geq 9$ & Confidence = 10 \\
\midrule
Command R                & 0.37 & 0.38 (99\%) & 0.49 (63\%) & 0.67 (31\%) \\
Command R+               & 0.36 & 0.36 (99\%) & 0.50 (63\%) & 0.66 (31\%) \\
Qwen-3 0.6B               & 0.00 & -0.00 (30\%) & 0.00 (18\%) & 0.00 (18\%) \\
Qwen-3 1.7B               & 0.16 & 0.08 (96\%) & 0.06 (89\%) & 0.00 (65\%) \\
Qwen-3 4B                 & 0.31 & 0.29 (90\%) & 0.15 (75\%) & 0.13 (41\%) \\
Qwen-3 8B                 & 0.33 & 0.33 (99\%) & 0.35 (81\%) & 0.33 (52\%) \\
Qwen-3 32B                & 0.35 & 0.37 (91\%) & 0.43 (63\%) & 0.60 (29\%) \\
Qwen-3 14B                & 0.34 & 0.36 (90\%) & 0.32 (63\%) & 0.40 (46\%) \\
Gemma-3 4B                & 0.20 & 0.20 (98\%) & 0.19 (93\%) & 0.24 (34\%) \\
Gemma-3 12B               & 0.03 & 0.03 (99\%) & 0.02 (98\%) & 0.02 (98\%) \\
Gemma-3 27B               & 0.02 & 0.02 (99\%) & 0.02 (99\%) & 0.02 (99\%) \\
LLaMA-3.2 1B    & 0.03 & 0.04 (71\%) & 0.02 (39\%) & -0.00 (12\%) \\
LLaMA-3.2 3B    & 0.16 & 0.14 (94\%) & 0.14 (82\%) & 0.11 (67\%) \\
LLaMA-3.1 8B    & 0.19 & 0.19 (99\%) & 0.26 (60\%) & 0.38 (39\%) \\
LLaMA-3.1 70B   & 0.22 & 0.22 (99\%) & 0.30 (56\%) & 0.26 (47\%) \\
\bottomrule
\end{tabular}
\caption{Open-weight judge Cohen's $\kappa$ when only considering predictions above varied confidence thresholds. Values in parentheses indicate the proportion of samples retained at each threshold. Some open-weight models are well-calibrated, with Command R reaching Cohen's $\kappa$ as high as 0.67 when giving a confidence score of 10.}
\label{appendix:table:confidence}
\end{table*}

\begin{table*}[h]
\centering
\begin{tabular}{lc}
\toprule
Writing Error Type & Cohen's $\kappa$ ($\downarrow$) \\
\midrule
no\_none\_of\_the\_above     & 0.9437 \\
no\_all\_of\_the\_above      & 0.8242 \\
ordered\_options            & 0.7379 \\
avoid\_k\_type              & 0.5921 \\
avoid\_negatives            & 0.5903 \\
no\_fill\_in\_blank         & 0.4706 \\
plausible\_distractors      & 0.4044 \\
no\_logical\_cues           & 0.3830 \\
problem\_in\_stem           & 0.3671 \\
grammatical\_consistency    & 0.3652 \\
no\_absolute\_terms         & 0.3576 \\
clear\_language             & 0.3496 \\
equal\_length\_options      & 0.3391 \\
focused\_stem               & 0.3117 \\
single\_best\_answer        & 0.2647 \\
no\_vague\_terms            & 0.2167 \\
avoid\_repetition           & 0.1602 \\
no\_extraneous\_info        & 0.1317 \\
no\_convergence\_cues       & 0.0904 \\
\bottomrule
\end{tabular}
\caption{Cohen's $\kappa$ breakdown for Cohere Command-R across writing error types. Lower values indicate weaker agreement. Command-R excels in simpler judgments like detecting certain options and option order, but disagrees with experts more on complex, subjective criteria like extraneousness, repetition, and convergence.}
\label{appendix:table:flaw_breakdown}
\end{table*}

%% file: appendix/prompts.tex
\hypersetup{
    linkcolor=white,
    citecolor=white,
    urlcolor=white
}

\lstset{
  literate={<}{{<}}1
           {>}{{>}}1
}

\begin{prompt}[title={Prompt \thetcbcounter: Multiple-Choice Question Answering Prompt (\cref{subsection:impact})}, label=prompt:mcqa]
Answer the following multiple choice question. The last line of your response should be of the following format: 'ANSWER: LETTER' (without quotes) where LETTER is one of \texttt{[ letters ]}. Think step by step before answering.\\

\texttt{[ question] }

\texttt{[ choices ]}
\end{prompt}

\begin{prompt}[title={Prompt \thetcbcounter: Contamination Detection (\cref{subsection:contamination})}, label=prompt:contamination]
You are an expert evaluator of dataset contamination for multiple-choice questions (MCQs). You are given a set of candidate source texts ("citations") and one MCQ. Your job is to determine whether the MCQ exists exactly, partially, or not at all in any of the citations.\\
<multiple-choice question>\\
Question: \texttt{[ question ]}\\
Correct Answer: \texttt{[ answer ]}\\
</multiple-choice question>\\

<citations>\\
\texttt{[ citations ]}\\
</citations>\\

Use the following criteria to determine the match type:\\
<matching criteria>\\
- "exact\_match": The question and correct answer appear verbatim or nearly verbatim in at least one of the citations\\ 
- "question\_match": The question appears verbatim or nearly verbatim in at least one of the citations, but not with the correct answer\\
- "partial\_match": It is possible to come up with the correct answer to the question based on information in the citations\\
- "no\_match": There is no information in the citations that can be used to answer the question\\
</matching criteria>\\

<general instructions>\\
- Use ONLY the information in the provided <citations>; ignore outside knowledge.\\
- When determining matches, do NOT consider punctuation or upper/lower casing.\\
- Check each citation independently.\\
- The citations index "i" is represented as <citation i></citation i>.\\
- Return every matching citation index in ascending order; if none match, return an empty list [].\\
- Provide a short, clear explanation for your decision, referencing the decisive overlaps when applicable.\\
</general instructions>\\

<format>\\
Return your output as valid JSON with the matching "result", the indexed "citations" that support your decision (empty list [] if "no\_match"), and an "explanation" for your decision:\\
\texttt{[ json format ]} \\
Do not include anything else.
</format>
\end{prompt}

\begin{prompt}[title={Prompt \thetcbcounter: Choices-Only Prompt (\cref{subsection:shortcuts})}, label=prompt:choices_only]
Answer the following multiple choice question just by using the choices and without access to the question. Use any strategy possible to come up with the correct answer, and then guess what the original/missing question was.\\

<choices>\\
\texttt{[choices]}\\
</choices>\\

<format>\\
Return your output as valid JSON with the key "answer" which is one of \texttt{letters}, "explanation" which is how you arrived at the correct answer, and "question" which is what you guess is the missing question.
\texttt{[json format]}\\
Do not include anything else.\\
</format>
\end{prompt}

\begin{prompt}[title={Prompt \thetcbcounter: Question Similarity Prompt (\cref{subsection:shortcuts})}, label=prompt:question_similarity]
You are an expert at determining whether a model was able to guess what the original multiple-choice question was just from the choices.\\

You will be given a multiple-choice question and the model's response. You need to determine whether the model was able to guess what the original question was just from the choices.\\

Here is the multiple-choice question:\\
<original question>\\
\texttt{[ question ]}\\
</original question>

Here is the model's response when answering just with the choices:\\
<response>\\
\texttt{[ response ]}\\
</response>\\

And the question that the model inferred\\
<inferred question>\\
\texttt{[ inferred question ]}\\
</inferred question>\\

To determine if the model successfully guessed the original question, use the following criteria:\\
- If the inferred question is an exact match or a semantic of the original question, return "exact\_match"\\
- If a test-taker who knew the answer to the inferred question would likely be able to answer the original question, return "knowledge\_match"\\
- In any other case, return "no\_match"\\

<format>\\
Return your output as valid JSON with the key "decision" which denotes the type of match between the inferred question and the original question.\\
\texttt{[ json format ]}\\
Do not include anything else.\\
</format>
\end{prompt}

\begin{prompt}[title={Prompt \thetcbcounter: Writing Flaw Prompt (\cref{subsection:writing_flaws})}, label=prompt:writing_flaw]
You are an expert evaluator of multiple-choice questions (MCQs). You are given the following writing rule:\\

<rule>\\
\texttt{[ rule ]}\\
</rule>\\

Your task is to judge whether a given MCQ **follows this rule**. Here are some guidelines for this specific rule:\\

<guidelines>\\
\texttt{[ guidelines] }\\
</guidelines>\\

<examples>\\
\texttt{[ examples] }\\
</examples>\\

<general instructions>\\
- Think carefully about whether the MCQ adheres to the rule.
- If the rule is clearly followed and there are no flaws in the MCQ, return "pass".\\
- If the rule is clearly violated and there are no flaws in the MCQ, return "fail".\\
- In borderline cases where you are unsure, return "pass".\\
- Provide a confidence score from 1-10 for your pass/fail decision - how strongly you believe the MCQ follows or violates the rule. 1 means not at all confident and 10 means very confident.\\
- Provide a short, clear explanation of your reasoning.\\
</general instructions>\\

Here is the MCQ to evaluate:\\
<multiple-choice question>\\
\texttt{[ mcq ] }\\
</multiple-choice question>\\

<format>\\
Return your output as valid JSON in the following format:\\
\texttt{[ format ] }\\
Do not include anything else.\\
</format>
\end{prompt}